\documentclass{article}
\usepackage[utf8]{inputenc}
\usepackage{color}
\usepackage[english]{babel}
\usepackage{array}
\usepackage{float}
\usepackage{mathtools}
\usepackage{url}
\usepackage{wrapfig}
\usepackage{bm}
\usepackage{multirow}
\usepackage{amsmath}
\usepackage{amsthm, amsfonts, soul}
\usepackage{amssymb}
\usepackage{graphicx}
\usepackage{subcaption}
\usepackage{float}
\usepackage{tablefootnote}
\usepackage{natbib}
\usepackage{adjustbox}
\usepackage{algorithm2e}

\usepackage[unicode=true,pdfusetitle,bookmarks=true,bookmarksnumbered=false,bookmarksopen=false,breaklinks=false,pdfborder={0 0 1},backref=false,colorlinks=false]{hyperref}

\usepackage[utf8]{inputenc} 
\usepackage[T1]{fontenc}    
\usepackage{hyperref}       
\usepackage{url}            
\usepackage{booktabs}       
\usepackage{amsfonts}       
\usepackage{nicefrac}       
\usepackage{microtype}      
\usepackage{xcolor}         
\usepackage{lastpage}
\usepackage{totcount}

\regtotcounter{startappendix}
\title{Efficient and Robust Bayesian Selection of Hyperparameters in  Dimension
Reduction for Visualization}

\author{%
  Yin-Ting Liao \\
 Department of Mathematics,
  University of California, Irvine \\
  Hengrui Luo\thanks{Corresponding author:  \texttt{hrluo@lbl.gov}} \\
  Lawrence Berkeley National Laboratory,
  \\ 
  Anna Ma\\
  Department of Mathematics, 
  University of California, Irvine
}

\begin{document}

\maketitle

\begin{abstract}
We introduce an efficient and robust  auto-tuning framework for hyperparameter selection in dimension reduction (DR) algorithms, focusing on large-scale datasets and arbitrary performance metrics. By leveraging Bayesian optimization (BO) with a surrogate model, our approach enables efficient hyperparameter selection with multi-objective trade-offs and allows us to perform data-driven sensitivity analysis.
By incorporating normalization and subsampling, the proposed framework demonstrates versatility and efficiency, as shown in applications to visualization techniques such as t-SNE and UMAP. We evaluate our results on various synthetic and real-world datasets using multiple quality metrics, providing a robust and efficient solution for hyperparameter selection in DR algorithms.
\end{abstract}

\section{Introduction}
 \subsection{Problem and Background}
The problem of data visualization in the big-data era is one of the central topics in modern data analysis and graphical statistics.  
As a crucial part of exploratory data analysis,  
it has been both a methodological and engineering challenge to visualize patterns in large-scale high-dimensional data. Recent research attempts have been extended into scalable methodologies for large-scale datasets with high-dimensionality \citep{wilkinson_distance-preserving_2022}, taking a generic regime consisting of dimension reduction (DR), subsampling, and sketching \citep{RandLAPACK_book}.   

One practical challenge in implementing DR methods for visualization on large datasets is hyperparameter selection. Novel DR methods, like those accommodating high-dimensional structures (e.g., \citet{luo2022spherical}, \citet{dover2022avida}), introduce more hyperparameters and in many cases, selection is crucial but expert knowledge is limited. For instance, 
\citet{szubert2019structure} used large neural networks for DR, adding hundreds of hyperparameters. 
Efficient hyperparameter selection is vital as it impacts reduced dataset quality and subsequent analyses. However, trying all configurations of hyperparameters (i.e., try-and-err) is unrealistic due to the computational cost of DRs.

We focus on hyperparameter selection for established DR methods for visualization and restrict to case studies using t-Stochastic Neighbor Embedding (t-SNE)~\citep{maaten2008visualizing} and Uniform Manifold Approximation by  Projection (UMAP)~\citep{mcinnes2018umap}.
However, as pointed out by \citet{coenen2022understand} and \citet{wattenberg2016use}, selecting appropriate hyperparameters is crucial in obtaining meaningful and interpretable results. 
As shown by recent works \citep{kobak2021initialization,damrich_contrastive_2022,bohm_attraction-repulsion_2022}, the choice of hyperparameters for t-SNE and UMAP impacts the quality of the data visualization.
Here, we briefly review t-SNE and UMAP and refer the reader to \cite{van2009learning} and \cite{mcinnes2018umap} for a more in-depth exposition of these methods.

\subsection{t-SNE and UMAP}
The t-SNE algorithm is
one of the most popular DR  techniques for visualization for high-dimensional data with many applications, including single-cell multi-omics \citep{kobak2019art}, bioinformatics \citep{li2017application}, and structural engineering \citep{hajibabaee2021empirical}. 
Informally, t-SNE finds a lower dimensional visualization of high dimensional data by preserving probability distributions over neighborhoods of points in high dimensions.

Until recently, data-dependent hyperparameter selection for t-SNE remains relatively unexplored  \citep{vu2021constraint}. The algorithm has two primary hyperparameters: perplexity, a proxy for the number of nearest neighbors, typically ranging from 5 to 50; and 
 an early exaggeration factor, affecting the tightness of natural clusters in visualization, with a default value of 12 in standard implementations.
The performance of t-SNE heavily depends on these hyperparameters.

UMAP \citep{mcinnes2018umap} is a versatile DR technique for visualization that is often used in bioinformatics \citep{becht2019dimensionality}, natural language processing \citep{pealat2021improved}, and image analysis\citep{allaoui2020considerably}. 
UMAP builds upon the idea that data points lie on a low-dimensional manifold embedded within the high-dimensional space. 
The algorithm employs Riemannian geometry and topological data analysis for manifold structure learning and lower-dimensional projection. 
 
Several hyperparameters influence UMAP's low-dimensional representation. We focus on optimally selecting the number of nearest neighbors and minimum distance. The number of neighbors balances local and global structure preservation, with smaller values emphasizing local structure. Minimum distance controls clustering compactness in lower-dimensional space, where smaller values create tighter clusters and larger values spread points. Typical choices for nearest neighbors and minimum distance are 2 to 100 and the interval $[0,1]$, respectively.

\subsection{Previous Work}

Previous works used manual selection, trial-and-error, and neural networks to tune hyperparameters for t-SNE and UMAP \citep{wattenberg2016use, du2022research, gove2022new, belkina2019automated, belkina2017visne,cao2017automatic}. Recent approaches have focused on tuning a single hyperparameter, such as exaggeration rate \citep{damrich_contrastive_2022,cai2021theoretical,bohm_attraction-repulsion_2022,kobak2021initialization}. In contrast, our framework efficiently handles multiple hyperparameters through Bayesian selection for DR techniques, providing a comprehensive and general approach.  

Most related to our work is the work of \citet{vu2021constraint}, where the authors devised $f_{score}$ for evaluating global quality for user-constrained DR method, along with Bayesian
optimization to optimize the score. Their investigation is limited to the $f_{score}$ metric, and requires  pairwise user constraints. Moreover, the performance of DR methods rely heavily on the initialization, and DR on large datasets (i.e., $n>3,000$) remain challenging to tune efficiently~\citep{bibal2019measuring}. Unlike \citep{vu2021constraint}, our approach incorporates subsampling and works for arbitrary metrics. This advantage of applicability for any metric offers a comprehensive, interpretable solution for DR hyperparameter selection with multi-objective trade-offs (for multi-objective tuning involving more than one performance metrics) and sensitivity analysis (for multivariate hyperparameters).  

Furthermore, by normalizing the parameter space \citep{xiao2022optimizing} and taking empirical averages (or quantiles) of repeated metrics from different intializations, we showed that the optimal hyperparameters learned from the subsample can be transferred back to the large datasets with good visualization. 

\section{Method Framework}

We store data as an $n\times d$ matrix $\bm{X}$, where $n$ is the sample size and $d$ is the dimensionality. 
We leverage subsampling to reduce $n$ to $n'\ll n$ and DR to reduce $d$ to $d'\ll d$, resulting in a reduced $n'\times d'$ matrix. Our goal is to minimize the loss of original structure when reducing size and dimensionality, making $n'$ and $d'$ suitable for tuning existing DR methods for visualization.

Besides expert evaluation, there are existing performance metrics that allows us to quantify the quality of DR results for visualization. Let $\mathcal{D}_{\gamma}:\mathbb{R}^{n\times d}\rightarrow\mathbb{R}^{n\times d'}$denote the mapping of an input dataset $\bm{X}$ to a lower dimensional representation via algorithm $\mathcal{D}$ that requires a pre-selected hyperparameter $\gamma$. For example, $\mathcal{D}$ can be the t-SNE algorithm and $\gamma$ the chosen perplexity. We denote the output as $\bm{X}^{*}=\mathcal{D}_{\gamma}(\bm{X})$. For visualization purposes \citep{wilkinson2012grammar}, we set $d'=2$ in the DR method but our frameworks works for any $d'$ and any metric in a scalable way.

\begin{wrapfigure}{l}[0pt]{0.5\textwidth} 
\includegraphics[width=0.5\textwidth]{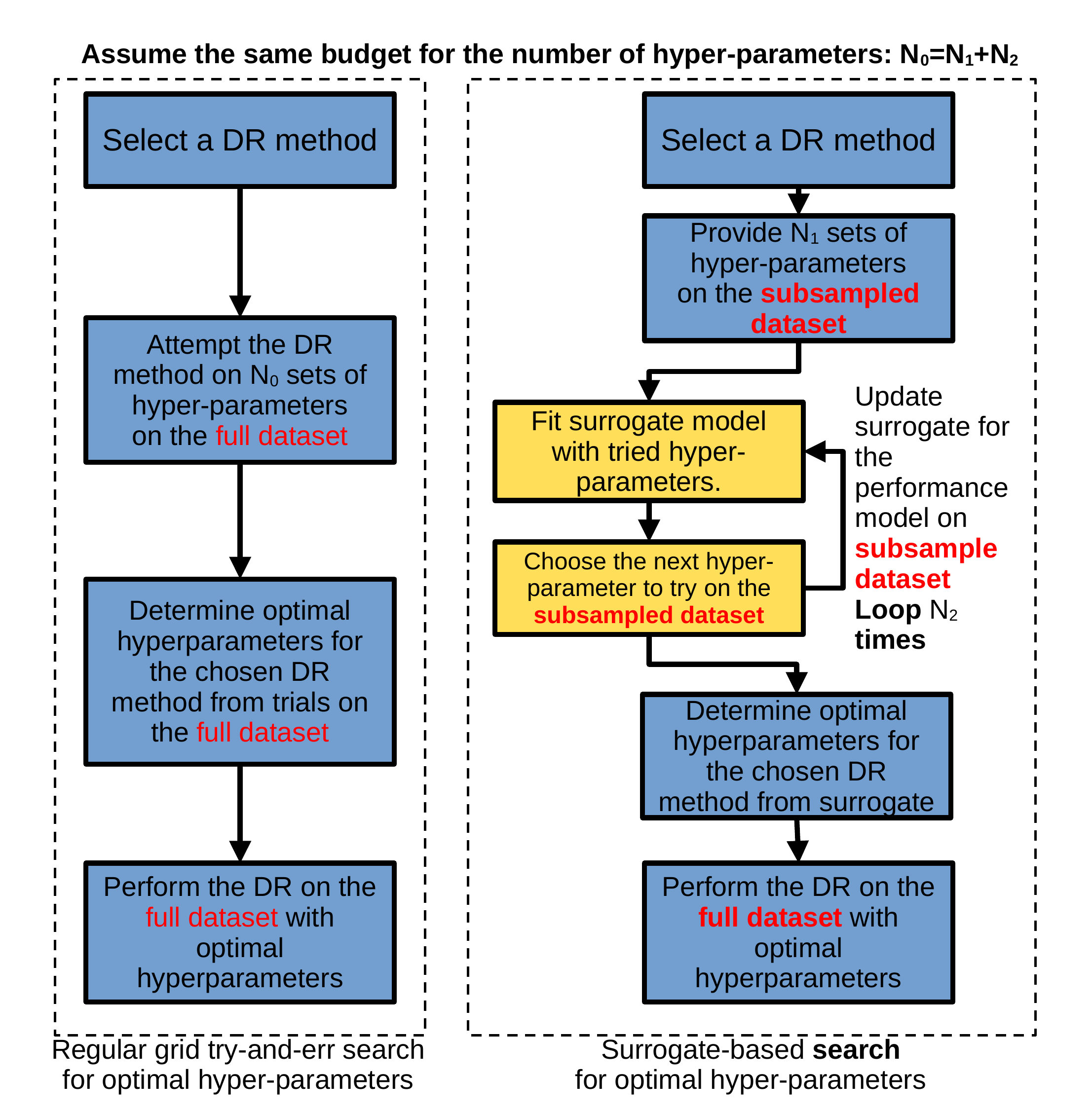}
\caption{\label{fig:Comparison-between-flowcharts}Comparison between flowcharts for (left) the try-and-err (or grid-search) and
(right) proposed surrogate-based search for optimal hyperparameters. Details are supplied in SM \ref{supp:algorithm}.
}
\end{wrapfigure}
We can compute a performance metric (e.g., coranking scores), $\tau:\mathbb{R}^{n\times d}\times\mathbb{R}^{n\times d'}\rightarrow\mathbb{R}$, between the original data $\bm{X}$ and processed data $\bm{X}^*$, where $\tau$ describes one aspect of the quality of the reduced data \citep{lee2007nonlinear}. The benefit of adopting a performance metric rather than expert evaluation is that we can quantify and model it as a function of its hyperparameters\footnote{Based on different definitions of performance metrics, metrics may have different ranges. We normalize each of these metrics to be in $[0,1]$ and enforce that smaller values of metrics indicate better DR results. }.

There are two categories of operational performance metrics we consider: task-independent and task-dependent performance metrics. As pointed out by \citet{lee2007nonlinear}, the primitive task-independent metrics will only depend on the original data matrix $\bm{X}$ and reduced data matrix $\bm{X}^{*}$. However, frequently we conduct subsequent analyses on reduced dataset $\bm{X}^{*}$, using task performance to characterize DR quality using task-dependent metrics.

After fixing the DR method $\mathcal{D}$, 
our problem of hyperparameter selection can be interpreted as optimizing a performance function $F(\gamma)\coloneqq\tau\left(\bm{X},\mathcal{D}_{\gamma}(\bm{X})\right)$ over a hyperparameter $\gamma$. While this formulation simplifies the problem, closed-form expressions of $F(\gamma)$ are not typically known, and classical (gradient-based) optimization methods are not applicable. To overcome this obstacle, we take the Bayesian optimization (BO) framework \citep{snoek2012practical,luo2021non,luo2022hybrid} select a surrogate model $\mathcal{S}$, and sequentially update the model to approximate $F(\gamma)$. 

In sharp contrast to previous work \citep{vu2021constraint}, we proposed to tune empirical means of several repeats  $\widehat{\mathbb{E}}F(\gamma)$ instead of $F(\gamma)$ directly due to the observation that metrics computed by DR methods are highly variable with different initializations \citep{coenen2022understand}. 
Intuitively, as the sequential samples of $(\gamma,F(\gamma))$ accumulate, the surrogate model targeting $\widehat{\mathbb{E}}F(\gamma)\approx\mathbb{E}F(\gamma)$ will be a better  approximation to $\mathbb{E}F(\gamma)$.
In conjunction with the heuristic sequential selection of candidate $\gamma$'s, we can obtain reasonable optimal $\hat{\gamma}$ that is sufficiently close to the true minimizer of $\mathbb{E}F(\gamma)$. This procedure is also referred to as \emph{surrogate-based hyperparameter tuning} in applications like machine learning \citep{snoek2012practical,shahriari2015taking}.

\subsection{Hyperparameter Tuning} 
We present a framework for tuning experiments for DR methods with single and multiple hyperparameters, outlined in Figure~\ref{fig:Comparison-between-flowcharts}, where we 
split the budget for pilot $N_{1}$ and sequential $N_{2}$ samples in our surrogate-based approach \footnote{We did not explicitly
specify a subsampling scheme in this flowchart for simplicity. We use \texttt{scikit-optimize} for basic experimental results, but also point out that \citet{HistoryDBUserGuide} provides a transfer-learning-oriented pipeline that records learned surrogates and supports the framework described in this paper.}.
Central trade-offs in our method are:
(1) Computational Efficiency: Balancing accuracy and efficiency hinges on surrogate model and optimization strategy choices, trading off between expense and optimization quality.
(2) Metric Robustness: Maximizing quality involves stable performance and mitigating randomness via averaging multiple-initialization metrics, measured by chosen metrics $\tau$, which further raises the computation complexity.

Averaging metrics is necessary to address the multiple possible sources of randomness in DR approaches. For example, one kind of randomness comes from the (random) initialization of the low-dimensional representation. Another type of randomness can be caused by subroutines within the algorithm (e.g., nearest neighbor search in UMAP). 
Given the fact we want to approximate  $\mathbb{E}F(\gamma)$ from multiple possible sources of randomness, we propose multiple-initialization averaged metrics for more stable and robust performance. In our experiments, we perform the DR for $N_{\text{repeat}}=10$ times using the same hyperparameter $\gamma$ and take the average metric to obtain $\widehat{\mathbb{E}}F(\gamma)$.

Our methodology consists of two main components: the objective function and the surrogate model $\mathcal{S}$. For the objective function, we need to match the parameter space of the reduced and the original dataset due to the lack of distance preservation in DR methods.  
There are several surrogate models  $\mathcal{S}$,  including the cheapest random selection, and the most expensive optimal benchmark, grid exhaustive search. We examine the following approaches, grid exhaustive search ($N_0=$ total sample  size), random forest regression, and Gaussian processes (GP-EI, GP-PI, GP-LCB \citep{louppe2017bayesian})
as Bayesian surrogates that can incorporate prior information with a total budget $N_0=N_1+N_2$, where $N_1$ is the budget for pilots and $N_2$ is the budget for the surrogate-based tuning. 
The sampling regime, normalization procedure and budget allocation are determining factors for surrogate quality. 

\subsection{Multi-objective Trade-offs}

Surrogate-based tuning faces challenges in balancing and optimizing competing objectives, especially as the number of goal metrics increases. This emphasizes the need for appropriate surrogate models and optimization strategies that efficiently handle multi-objective problems, balancing accuracy and computational efficiency. Rather than focusing on optimizing metrics, we construct sensitivity analyses based on surrogates to detect sensitive  hyperparameters (SM \ref{sec:SA}). 
With multiple goal metrics (e.g., AUC and Q-global in Figure \ref{fig:unbalanced_metric}), conventional single-objective optimization techniques are insufficient due to conflicting objectives. We propose the adoption of multi-objective optimization algorithms, such as Pareto-based approaches, to identify non-dominated solutions representing optimal trade-offs.

Pareto fronts \citep{miettinen1999nonlinear} provide non-dominated solutions, balancing conflicting metrics. Multi-objective optimization algorithms, such as NSGA-II \citep{deb2002fast}, efficiently search for Pareto optimal solutions. In our approach, we construct the Pareto front using samples from the surrogate model, noting that it can also be elicited directly from the fitted surrogates for two metrics $F_1(\gamma),F_2(\gamma)$ where the perplexity $\gamma$ could affect both metrics simultaneously. After obtaining the Pareto front, decision-makers analyze trade-offs and select suitable solutions based on requirements or preferences, possibly choosing a point with the best balance or using additional criteria like the knee point.

\section{Numerical Experiments}

\subsection{Datasets}
\label{sec:data}
We consider the following four synthetic and real-world datasets, and provide experiments on additional datasets in SM \ref{supp:sine dataset} and \ref{supp:vehicle dataset}.  The \textbf{Unbalanced 2-cluster dataset} is a small-scale, synthetic dataset that captures the performance of our framework when applied to a small class unbalanced dataset. It consists of 60 points, 10 in one class and 50 in another class. The \textbf{COIL}~\citep{coil20} is a real-world dataset of images of physical objects taken from different angles. We take $n=360$ from 5 classes in this data. Lastly, we consider two real-world, large-scale datasets. \textbf{Reuters-English}~\citep{amini2009learning} for which we use $n\approx 14,000$ data points and \textbf{Single cell transcriptomics} with $n \approx 23,000$~\citep{tasic2016adult}. These real-world datasets exceed the maximum $n\approx3,000$ in previous work \citep{vu2021constraint}.

Using the unbalanced 2-cluster dataset, we illustrate how the surrogate-based tuning methodology works when the evaluation metrics can be directly computed on the full dataset. We show that our approach can find a set of hyperparameters that optimize the chosen metrics. 
Using the COIL dataset, we demonstrate that subsampling will speed up the tuning process without significantly impacting the outcome of hyperparameter optimization. 
The trend (hence the optimal hyperparameters) of each metric computed on subsamples is preserved when compared to the metric computed on the full dataset at the cost of increased uncertainty. Lastly, we apply our entire pipeline of the subsample-and-tune method and carry out corresponding analyses on the Reuters-English dataset and the single cell transcriptomics dataset to demonstrate efficacy on large-scale data.

\subsection{Evaluation Metrics}
We employ multiple quality metrics to evaluate t-SNE and UMAP.
The metrics we consider fall into two categories: task-independent and task-dependent. Task-independent metrics describe data properties, e.g., distance preservation, while task-dependent metrics focus on downstream tasks' performance, reflecting real-world utility. Combining both types enables a comprehensive evaluation, emphasizing strengths and weaknesses of DR techniques \citep{bibal2019measuring}.

For \emph{task-independent} metrics, we investigate Q-global, average pairwise distance ratio, and Q-local metrics from \citet{lee2009quality} and Area Under the Curve (AUC) from~\citep{lee2015multi}. Q-global and average pairwise distance ratio (between the pairwise distances of the original and reduced dataset) evaluate distributional pairwise distance preservation, while Q-local focuses on local neighborhood structures. These metrics determine DR effectiveness in preserving data relationships and are used to assess DR quality.
The AUC metric is also used to evaluate the quality of DR techniques. The AUC quantifies the area under their proposed functional metric $R_{NX}(K)$,
curve (varying scale $K$), providing a scalar-valued metric that allows comparison of the average (across all possible scales) quality of the resulting DR embedding compared to a random embedding. A higher AUC indicates better performance of the model.
For \emph{task-dependent} metrics, we evaluate subsequent task performance on reduced datasets, expecting minimal performance loss if the DR method preserves essential information. Normalized Mutual Information (NMI, \citep{strehl2002cluster}) evaluates data structure preservation and is suitable for unsupervised tasks. For supervised tasks like classification, we assess misclassification rates using a simple \emph{logistic regression} model \citep{mccullagh2019generalized}. The classifier uses the 80\% of the data for training and the 20\% for testing, enabling DR method comparisons by evaluating the preservation of the original classification structure.

\subsection{Numerical Results} 

In this section, we provide empirical results using our framework to choose t-SNE's perplexity parameters on the four datasets discussed in Subsection~\ref{sec:data}. To ensure the parameter spaces are the same for full and subsampled datasets and facilitate the comparison of perplexities between the original and subsampled data, we use \emph{normalized perplexity}, calculated by dividing the perplexity by the number of data points. This rough perplexity normalization is intuitively consistent with more principled perplexity normalization using asymptotics \citep{xiao2022optimizing}. We perform each surrogate-based tuning for 20 batches and provide the budget for pilot $N_1=5$ and the budget for tuning $N_2\in\{15,20\}$ depending on the tasks. 

We also include experiments on tuning for UMAP focusing on two hyperparameters -- the number of nearest neighbors (n\_neighbor), between 2 and 100 as suggested in the UMAP documentation, and minimum distance (min\_distance), in the interval [0,1]. Due to this larger dimension of the hyperparameter space, we assign a higher budget $N_2\in\{20,40\}$. In order to account for the difference in sizes between the original and subsampled data, we use a normalized n\_neighbor, which is similar to the normalized perplexity used for t-SNE.
For the real-world dataset, we reduce the parameter space of min\_dist from the interval $[0,1]$ to points $\{0,0.2,0.4,0.6,0.8,1 \}$, since the sensitivity of min\_dist is smaller compared to that of n\_neighbor as is observed in SM \ref{sec:SA}.

\begin{wrapfigure}{l}{0.5\textwidth} 
\includegraphics[width=.5\textwidth]{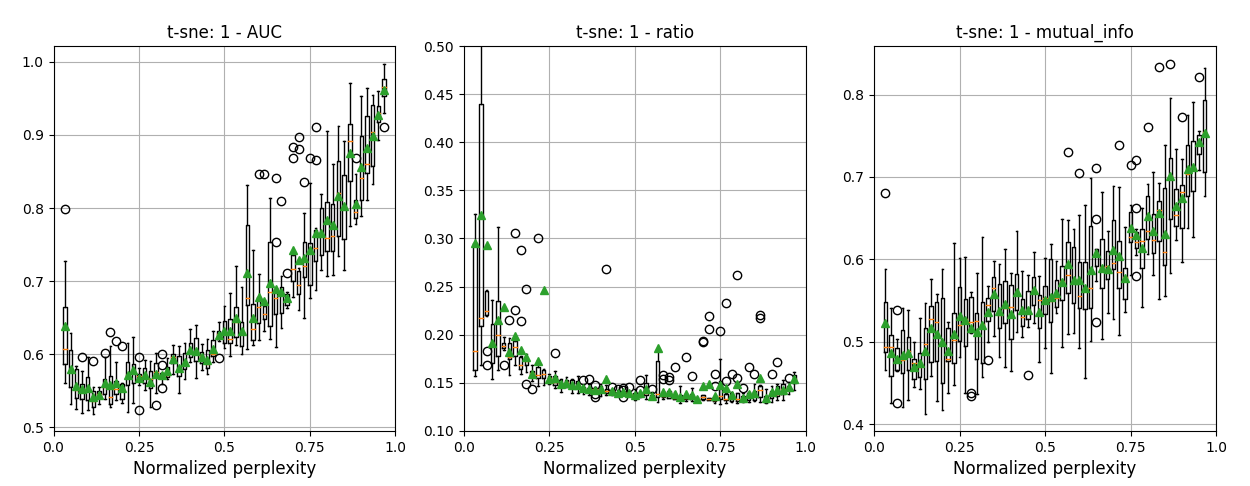}
 \caption{
 Boxplots for the performance of t-SNE using AUC, average ratio, and NMI for all possible  normalized perplexities, where the medians are marked by green triangles.
 }
 \vspace{-0.5cm}
    \label{fig:unbalanced_metric}
\end{wrapfigure}

\textbf{Example 1. Unbalanced 2-cluster dataset.}
In our first experiment, we demonstrate our framework on parameter selection for t-SNE applied to the unbalanced 2-cluster dataset. 
Since perplexity serves as a proxy for the number of nearest neighbors, we intuitively expect to obtain optimal results when the perplexity does not exceed the size of a single cluster.
We focus on three performance metrics: AUC, average ratio and NMI, to analyze the influence of t-SNE's initialization on the resulting lower-dimensional dataset.

We begin by considering a simple grid search over all possible normalized perplexity values. Our experiments suggest that most metrics are sensitive to initialization, and averaging the results over multiple initializations can help stabilize the outcome. 
We demonstrate this phenomenon with the 2-cluster dataset, and the results are shown in Figure~\ref{fig:unbalanced_metric}.
\begin{wrapfigure}{l}{0.5\textwidth}
\includegraphics[width=0.5\textwidth]{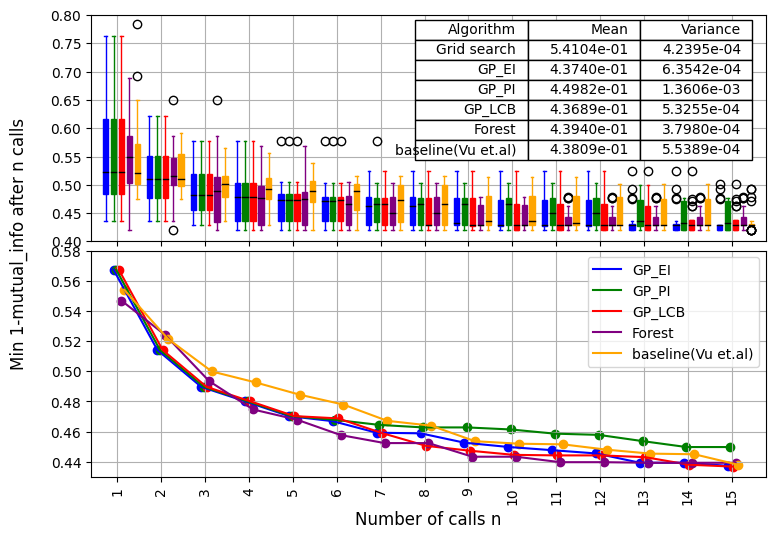}
  \caption{Comparison between convergence plots for surrogate-based tuning using GP-EI/GP-PI/GP-LCB/forest and \citep{vu2021constraint} for the $2$-cluster dataset.}
  \label{fig:unbalanced}
\end{wrapfigure} 

We observe that $1-\text{AUC}$ is minimized when the normalized perplexity is approximately 0.15 (perplexity is $9 = 0.15 \times 60$). The clustering performance under t-SNE, as evaluated by NMI, is also best when the normalized perplexity is around $0.15$. For unbalanced datasets, based on a grid search the optimized normalized perplexity is $0.16$, where the performance is optimal in terms of both the AUC  (task-independent) and NMI (task-dependent). Performance is not as favorable when the perplexity falls below or exceeds this value. On the other hand, given the average ratio (task-independent), the optimal normalized perplexity is at 0.75 (or perplexity $45 =0.75 \times 60$).

\begin{wrapfigure}{r}{0.5\textwidth}
  \includegraphics[width=0.5\textwidth]{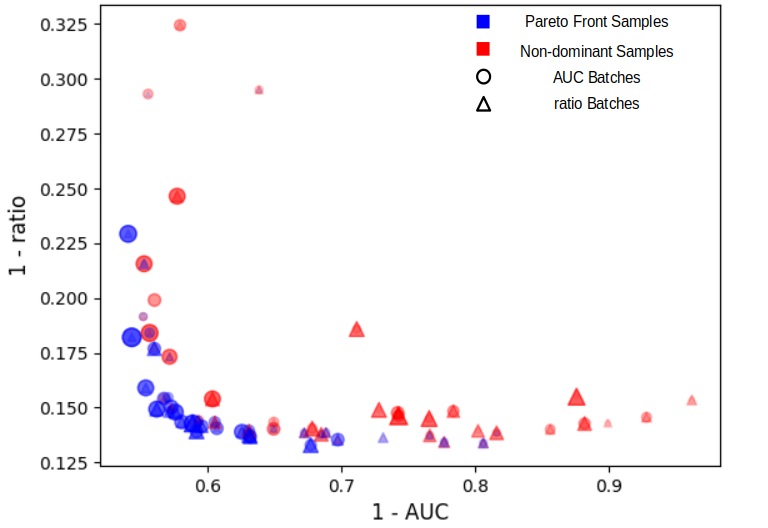}
  \caption{Blue represents the Pareto front and red represents the rest
sequential samples. Marker size correlates with the number of tuning samples (i.e., repeats during tuning) for the 2-cluster dataset.}
  \label{fig:unbalanced_pareto_history}
\end{wrapfigure}

Next, we use the NMI for surrogate-based perplexity tuning.
The result is given in Figure \ref{fig:unbalanced}. Among the $20$ batches, the minimum of $1 - \text{NMI}$ converges within $15$ iterations. In grid search, we use the budget of $N_0=60$ 
and obtain the minimal $1 - \text{NMI}$ of $0.541$, whereas using the surrogate-based tuning obtained a minimal $1- \text{NMI}$ of $0.537$, using a lower budget of $N_0 = 20$. We compare our results with \citep{vu2021constraint} and observe that our approach effectively reduces the variance and enhances the robustness of the tuning process, particularly when the metric under consideration are sensitive to initialization (e.g., AUC and NMI in Figure \ref{fig:unbalanced_metric}).
Finally, we include Pareto fronts using samples from the surrogate model. Figure \ref{fig:unbalanced_pareto_history} includes samples from tuning samples based on AUC and average ratio. 
We include both samples to balance the two tuning processes, as those circle markers based on AUC appear mostly on the left.
We observe that the Pareto fronts for AUC versus average ratio form a surface rather than a singular point. Hence, to address more than one objective and account for the trade-off between different metrics, our method recommends 
blue points in Figure~\ref{fig:unbalanced_pareto_history}.

\textbf{Example 2. COIL dataset.}
In this experiment, we demonstrate how subsampling techniques effectively enhance the efficiency of DR methods. 
We subsample a third of the original dataset under two different sampling schemes: random sampling and leverage score sampling \citep{ma2014statistical}.  
Figure \ref{fig:coil20-subsample} shows we obtain similar results for the two sampling schemes (See also SM \ref{supp:sine dataset}). Under leverage score sampling, the variation for the performance metrics is smaller compared to that under random sampling.
Moreover, using the subsampled dataset, the landscape of the performance is similar to the original dataset. Thus, to find the optimal perplexity for the full dataset, it suffices to obtain the optimal normalized perplexity for the subsampled dataset.
 In Figure \ref{fig:coil20_full}, we see that the classification performs relatively well when the perplexity is small. While the AUC is minimized when the normalized perplexity is around $0.12$, the Q-global is minimized at 0.9. We plot the visualization corresponding to these two optimal perplexities in SM \ref{SM:Visual_COIL}.
 
\begin{wrapfigure}{l}{0.55\textwidth}
    \includegraphics[width=0.55\textwidth]{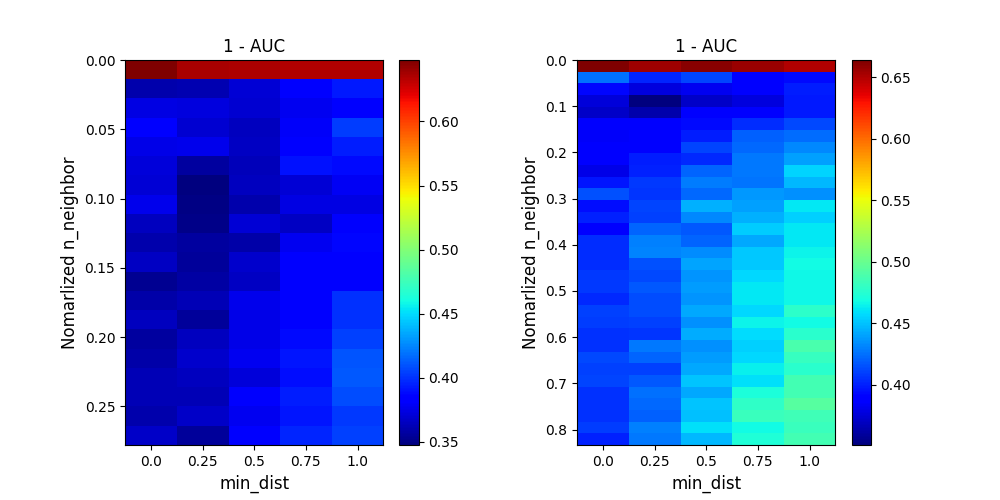}
  \caption{Heatmaps of mean metrics  for UMAP on full data (left) and subsampled data (right) (variance is bounded by 0.001) from COIL-20}
  \label{fig:coil5_heatmap}
\end{wrapfigure}
\begin{figure}[t]
\centering
\adjustbox{max width=1.2\textwidth}{
\includegraphics[width=1.0\textwidth]{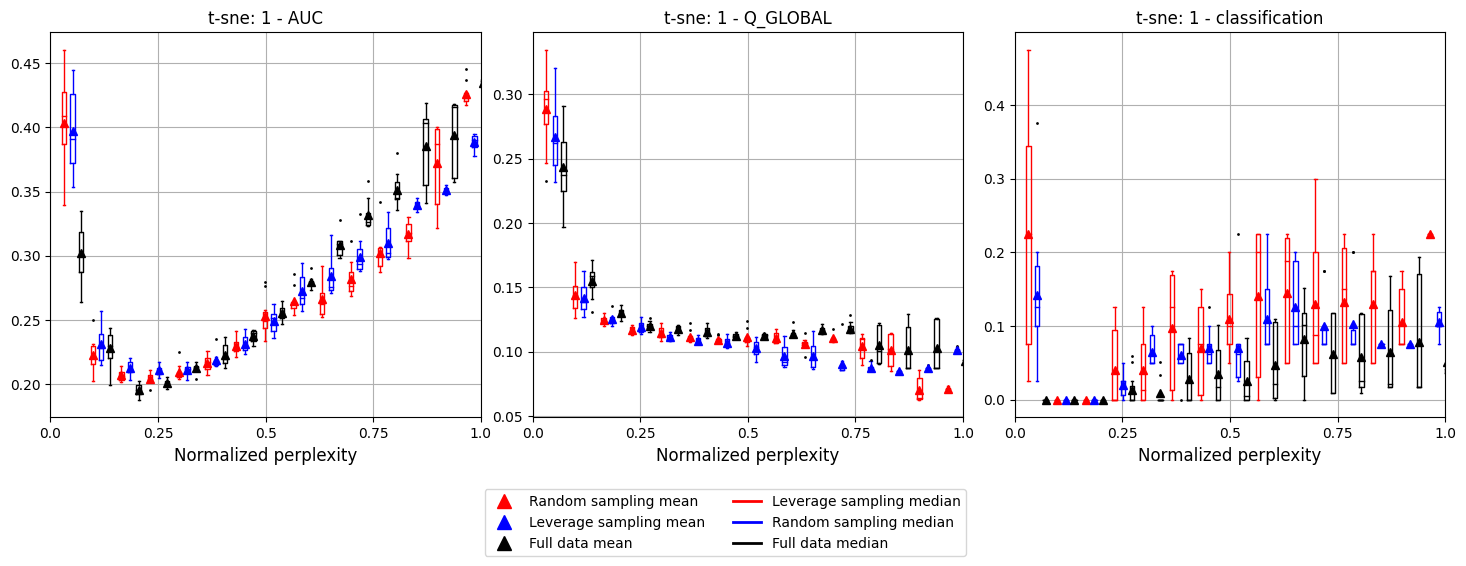}
}
 \caption{Comparison between full data and different sampling schemes for t-SNE on COIL dataset with the boxes denoting the median and 25\%-75\% quantiles (we use sparse perplexities sequence). }\label{fig:coil20-subsample}\label{fig:coil20_full}
\end{figure}

\begin{figure}[t]
\begin{subfigure}[b]{0.45\textwidth}
\includegraphics[width=\linewidth]{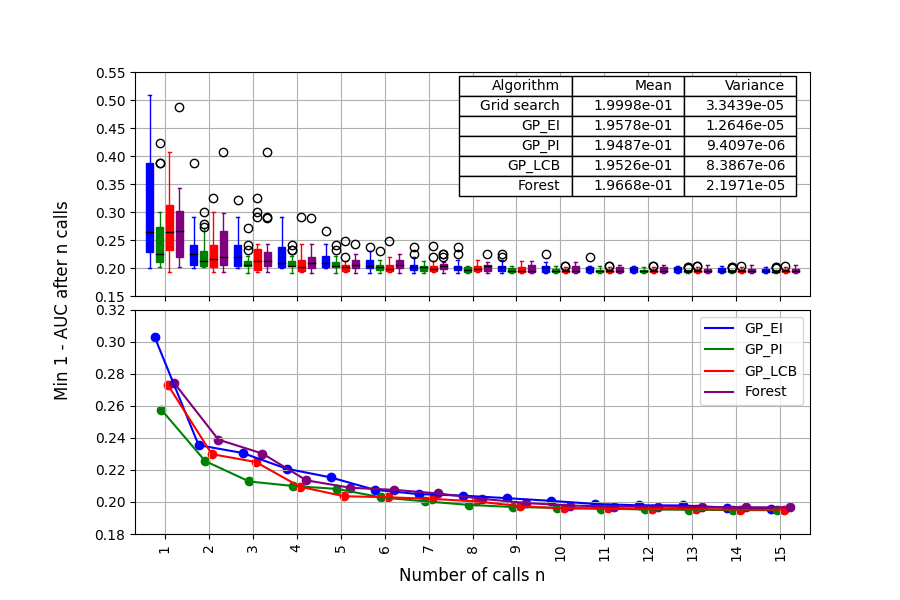}
\caption{Tuning for t-SNE}\label{fig:coil20_cvgplot}
\end{subfigure}  
\begin{subfigure}[b]{0.45\textwidth}
\includegraphics[width=\linewidth]{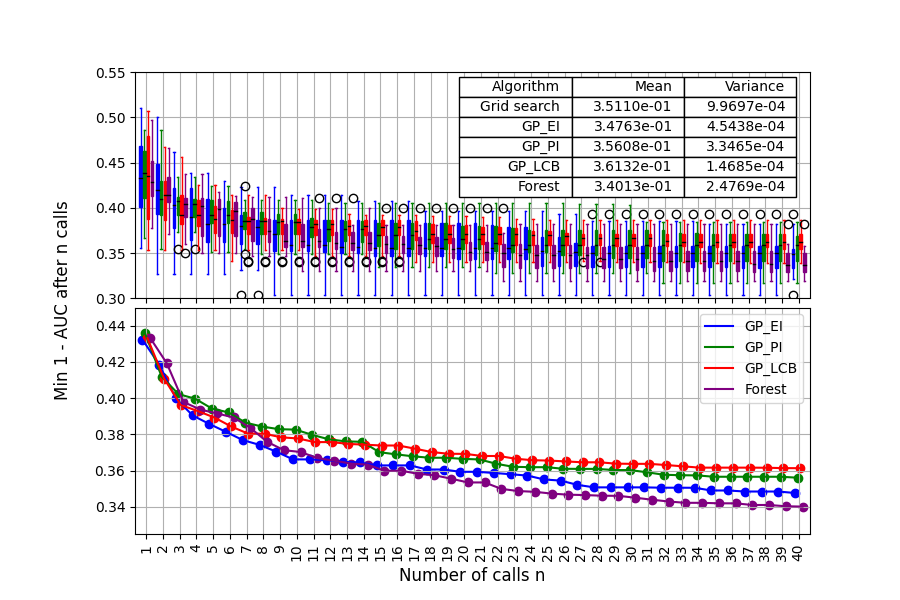}
\caption{Tuning for UMAP} 
\label{fig:coil5_gp_subsample}
\end{subfigure}
 \caption{Comparison between convergence plots for surrogate-based tuning using GP-EI/GP-PI/GP-LCB/forest  for tuning on the subsampled COIL dataset.
 } \label{fig:coil20-gp}
\end{figure}

Finally, we run the surrogate-based tuning for the subsampled data to find the optimal perplexity according to AUC and assign the budget of $N_0=20$ (as opposed to a sparse grid search, which requires the budget $N_0=60$). The result is shown in Figure \ref{fig:coil20-gp}. We see that within $5$ iterations, the average  converges to the minimum $1-$ AUC. 
This experiment demonstrates that the subsampled data is a good proxy for the original data in tuning for optimal perplexity. The surrogate-based tuning in the subsampled data is efficient and accurate.

Our framework is not limited to one hyperparameter or a single DR approach. We present how subsampling techniques enhance the efficiency of tuning UMAP.  We plot the heatmaps for the mean $1-\text{AUC}$ in Figure \ref{fig:coil5_heatmap} and see that the optimal $1-\text{AUC}$ is obtained at normalized n\_neighbor $0.1$ and min\_distance $0.25$ for both the subsampled and full dataset. This justifies the use of a subsampled dataset as a proxy of the full dataset when tuning hyperparameters in UMAP. Finally, we run the surrogate-based tuning using the subsampled data, as shown in Figure \ref{fig:coil5_gp_subsample}. Amongst the 20 batches, we obtain the minimum mean $1-\text{AUC}$ of $0.3414$, which is comparable to that of $0.3511$ for a sparse grid search with budget $N_0=150$.

\begin{figure}[t]

\begin{subfigure}[b]{0.45\textwidth}
\includegraphics[width=\linewidth]{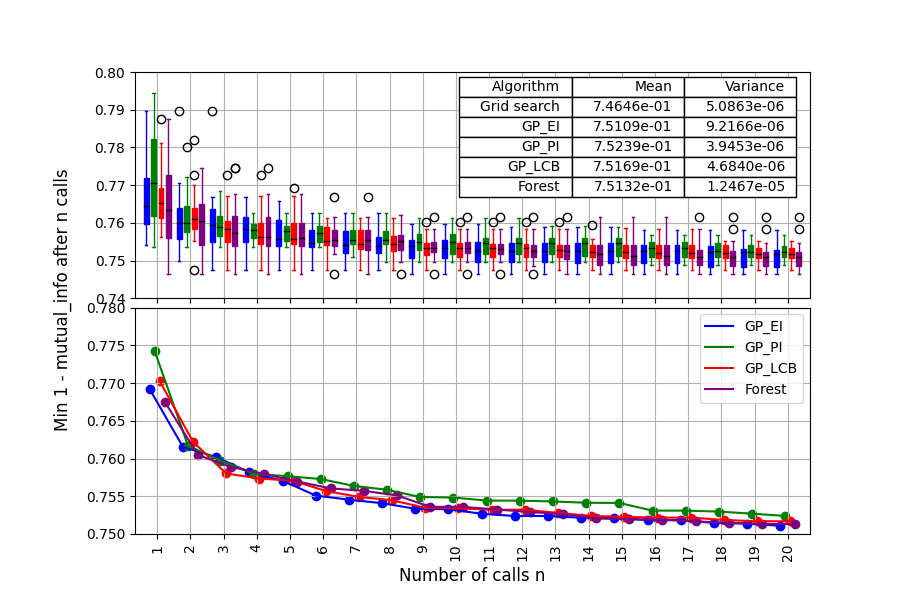}
\caption{Tuning for t-SNE}
\label{fig:reuters-tsne}
\end{subfigure}  
\begin{subfigure}[b]{0.45\textwidth}
\includegraphics[width=\linewidth]{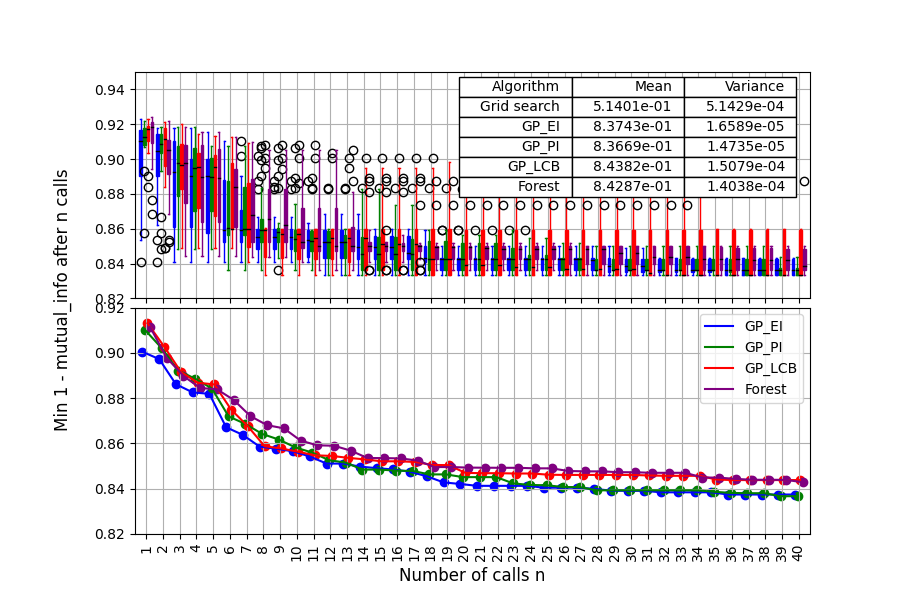}
\caption{Tuning for UMAP}
\label{fig:reuters-umap}
\end{subfigure}
\caption{Convergence plot for surrogate-based tuning for the Reuters English dataset.
 } \label{fig:reuters}
\end{figure}

\begin{figure}[t]

\centering
\begin{subfigure}[t]{0.3\textwidth}
\centering
\includegraphics[width=\linewidth]{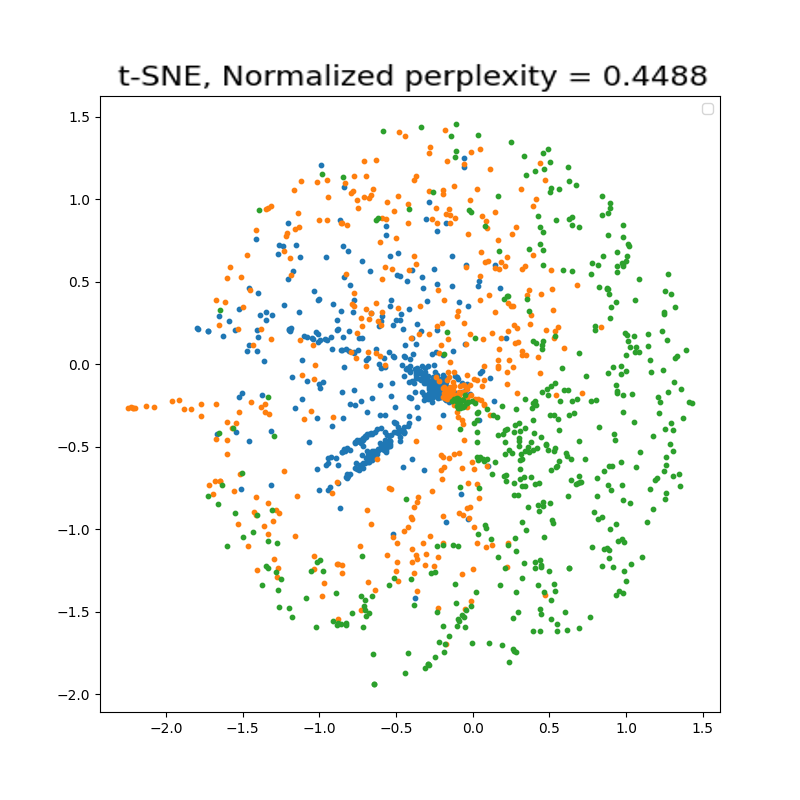}
\end{subfigure}
\begin{subfigure}[t]{0.3\textwidth}
\includegraphics[width=\linewidth]{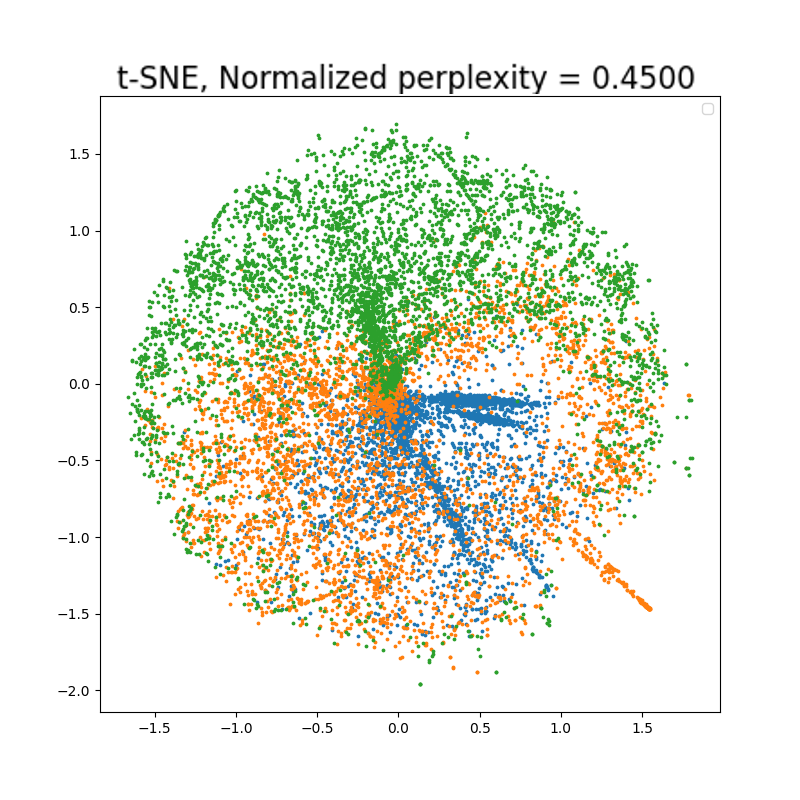}
\end{subfigure}
\begin{subfigure}[t]{0.3\textwidth}
\includegraphics[width=\linewidth]{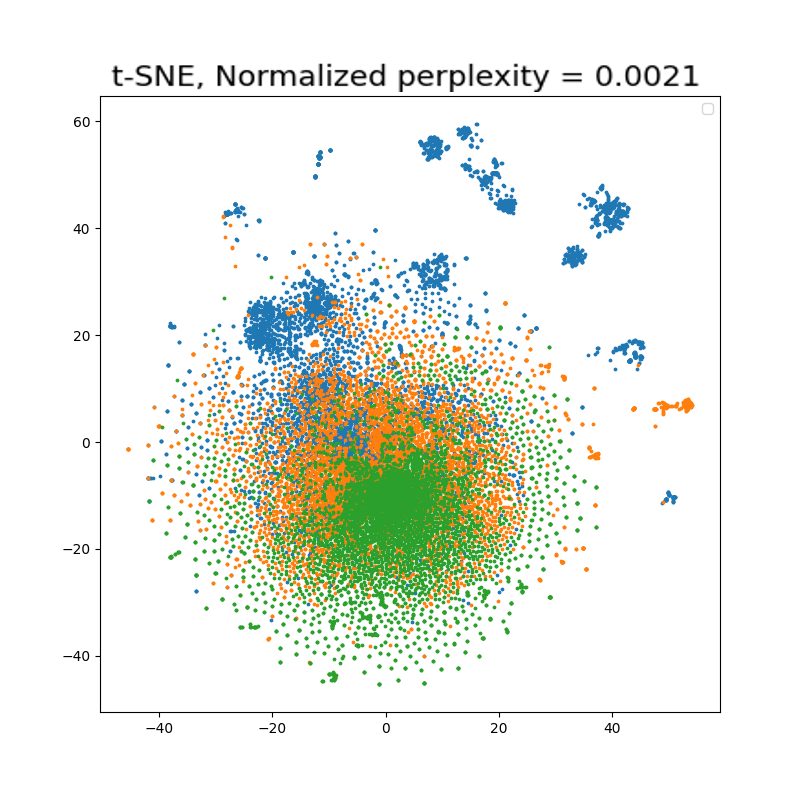}
\end{subfigure}
 \caption{Visualization of the subsampled data (left) and full data (middle) at the optimal normalized perplexity, and
 the full data for the default perplexity, 30 (right).
 }\label{fig:reuters-visual}
\end{figure}

\textbf{Example 3. Reuters English dataset.}
This experiment demonstrates the subsample-and-tune pipeline, which overcomes the sample size bottleneck in the simple BO method without subsampling by \citet{vu2021constraint}. We subsample a tenth of the full dataset, obtain 1,426 data points using random sampling, and run the surrogate-based tuning for t-SNE with the performance metric NMI. The results for different surrogates are plotted in Figure~\ref{fig:reuters-tsne}. The budget for grid search is $N_0=1,426$ while the budget for the surrogate-based tuning is $N_0=25$. We observe that for all 4 kinds of surrogates, the tuning processes converge to the minimal $1-\text{NMI}$ within 10 iterations. Moreover, the optimal normalized perplexities obtained from grid search and surrogate-based tuning are 0.47 and 0.45, respectively. We next take the normalized perplexity from tuning using subsampled data and obtain the t-SNE visualization of the full data in Figure \ref{fig:reuters-visual}. We observe that at the same normalized perplexity, the visualizations of the subsampled and full data are similar. Moreover, we compare the optimal normalized perplexity with default perplexity. Since the optimum is tuned under NMI, we see that the clustering result in the optimal perplexity is better than that for the default perplexity and leads to a better t-SNE visualization using fine-tuned perplexity as shown in Figure 2(f) in \citet{liu2022stationary}.

In addition, we present the convergence plot for the subsample-and-tune pipeline when applied to UMAP in Figure \ref{fig:reuters-umap}.  The tuning processes for different surrogates, $N_0=45$ converge within 10 iterations and the optimal values obtained is closed  to the grid search, which uses the budget $N_0=600$.

\begin{figure}[t]

\begin{subfigure}[b]{0.45\textwidth}
\includegraphics[width=\linewidth]{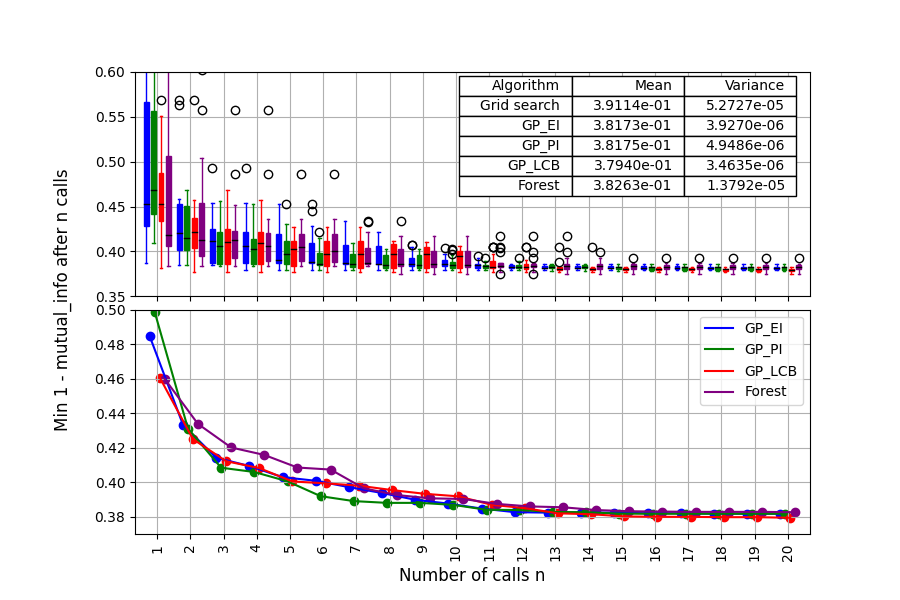}
\caption{Tuning for t-SNE}
\end{subfigure}  
\begin{subfigure}[b]{0.45\textwidth}
\includegraphics[width=\linewidth]{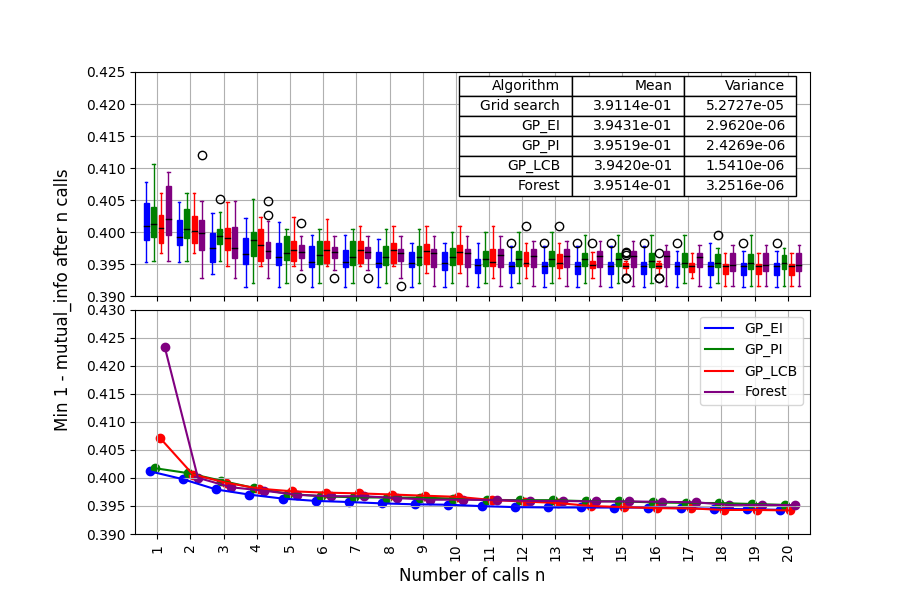}
\caption{Tuning for UMAP}
\end{subfigure}
\caption{Convergence plot for surrogate-based tuning for the single cell dataset.
 } \label{fig:tasic}
\end{figure}

\begin{figure}[t]

\centering
\begin{subfigure}[t]{0.3\textwidth}
\centering
\includegraphics[width=\linewidth]{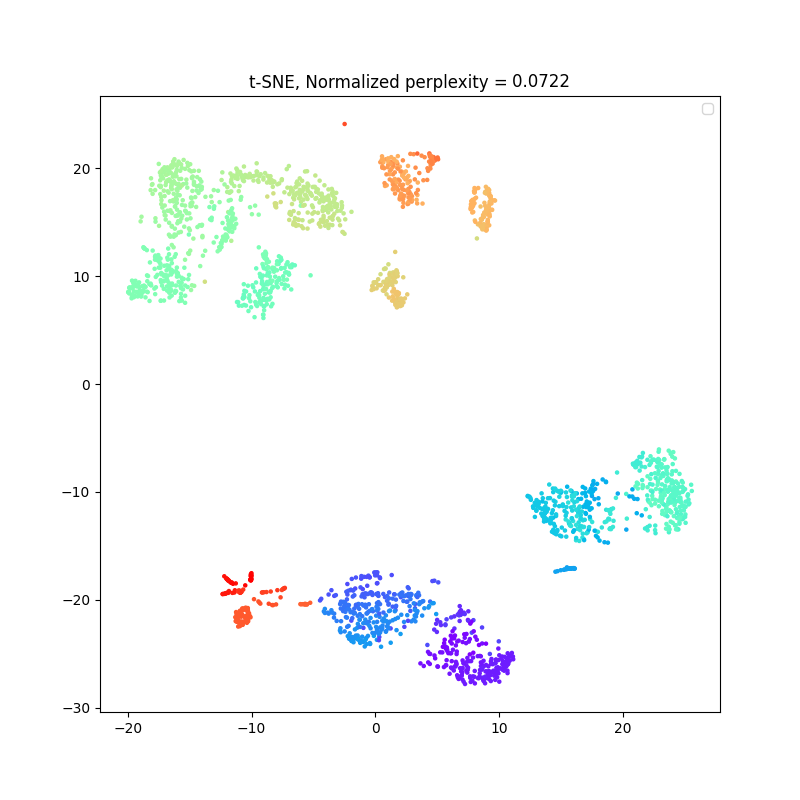}
\end{subfigure}
\begin{subfigure}[t]{0.3\textwidth}
\includegraphics[width=\linewidth]{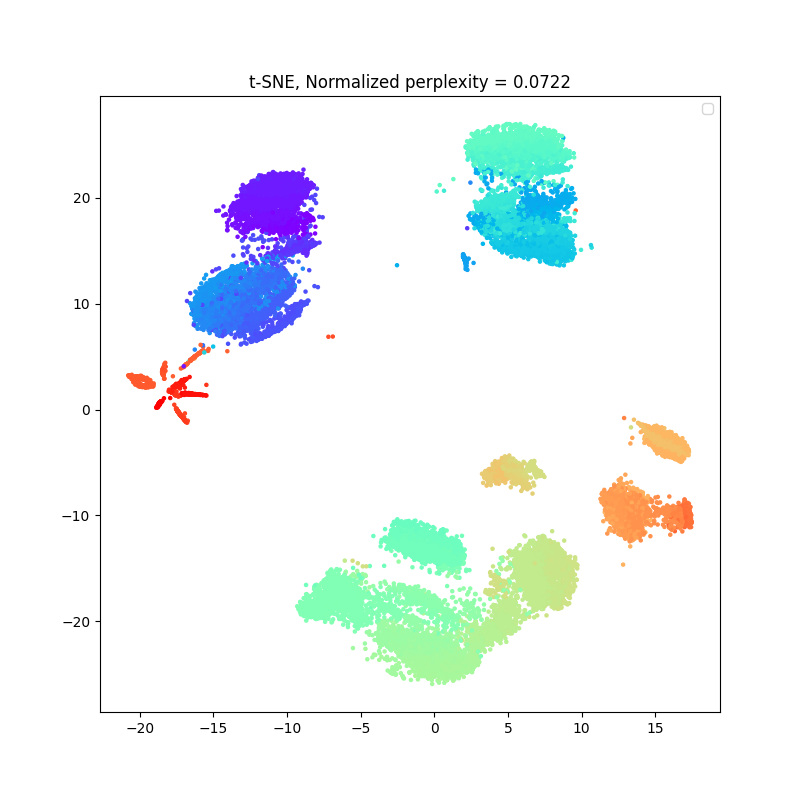}
\end{subfigure}
\begin{subfigure}[t]{0.3\textwidth}
\includegraphics[width=\linewidth]{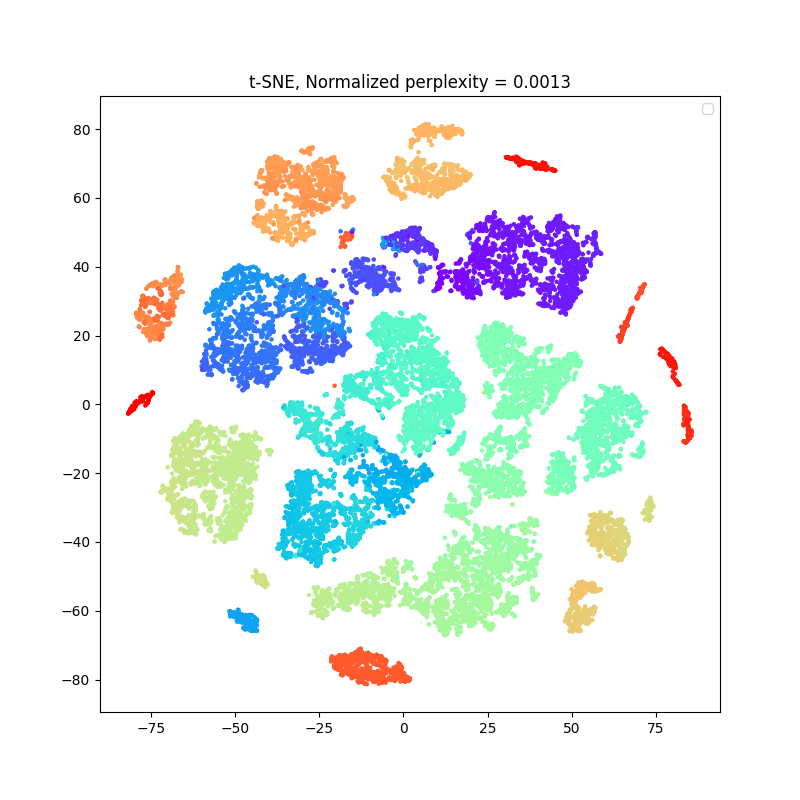}
\end{subfigure}
\begin{subfigure}[b]{0.3\textwidth}
\centering
\includegraphics[width=\linewidth]{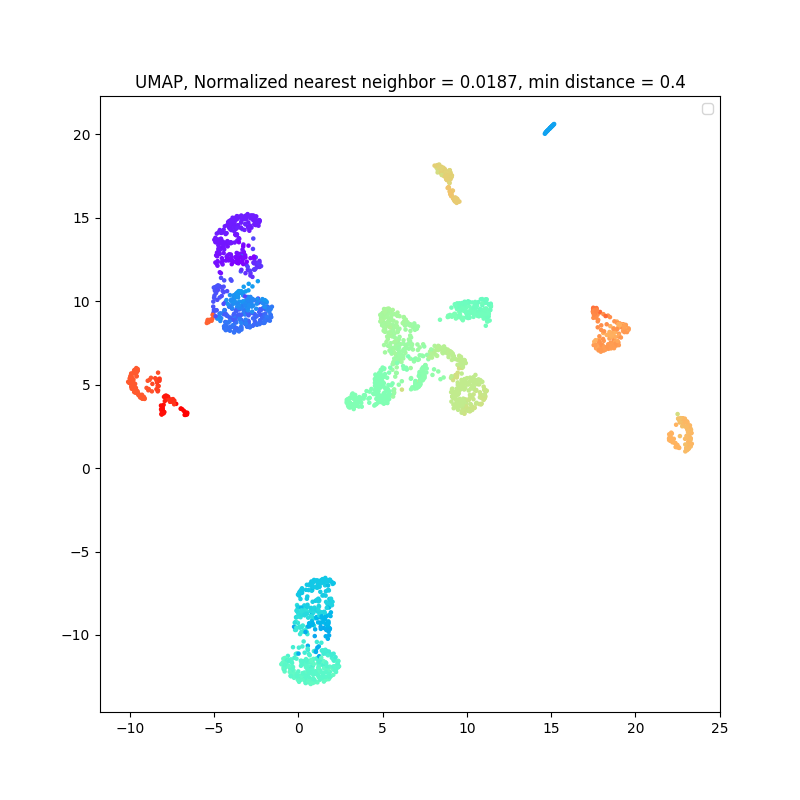}
\end{subfigure}
\begin{subfigure}[b]{0.3\textwidth}
\includegraphics[width=\linewidth]{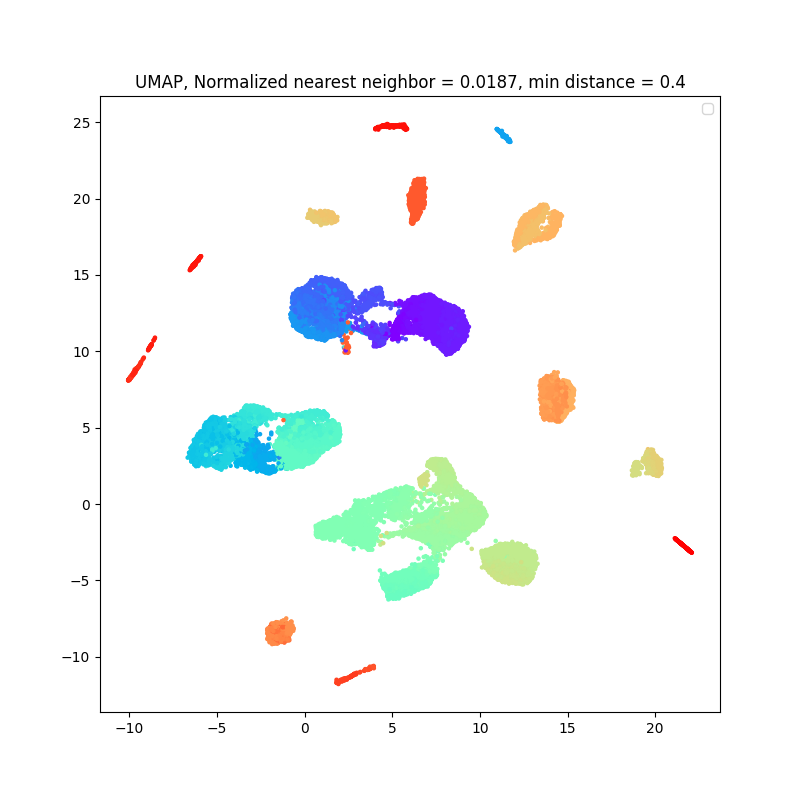}
\end{subfigure}
\begin{subfigure}[b]{0.3\textwidth}
\includegraphics[width=\linewidth]{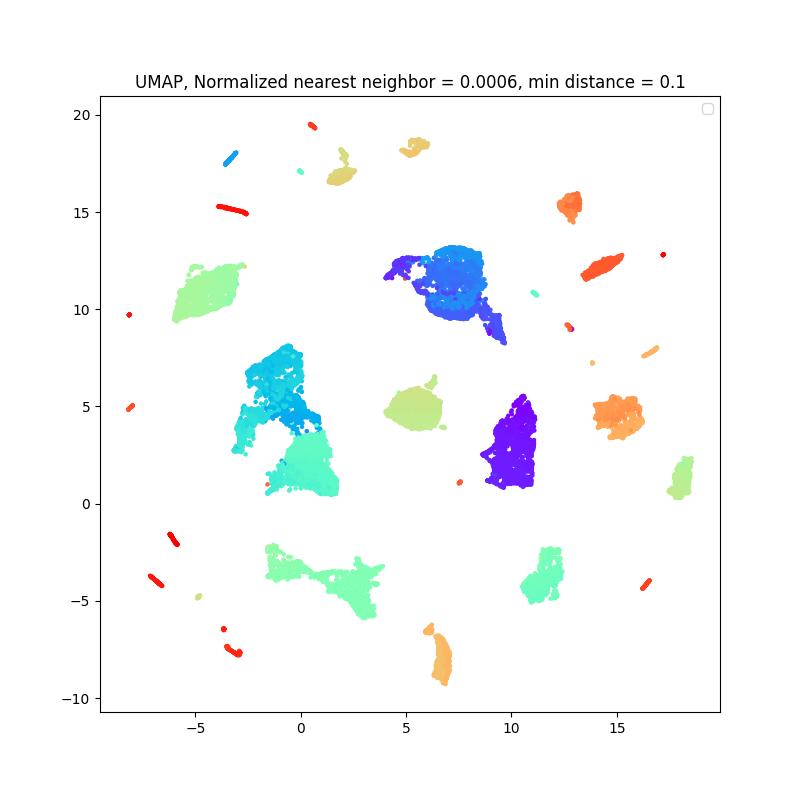}
\end{subfigure}
 \caption{Visualization using t-SNE (top row) and UMAP (bottom row) of the subsampled data (left) and full data (middle) at optimal hyperparameters, and
 the full data for default parameters (right).
 }\label{fig:tasic-visual}
\end{figure}

\textbf{Example 4. Single cell transcriptomics dataset.}
We subsample a tenth from the full data and obtain 2,282 data and perform surrogate-based tuning for both t-SNE and UMAP using NMI on the subsampled data. The results are shown in Figure \ref{fig:tasic}. With much lower budgets, $N_0=25$, we obtain the optimum values comparable to those found using grid search, $N_0=2,280$ and 600 for t-SNE and UMAP, respectively. We next use the obtained hyperparameters to recover the 2-dimensional visualisations for the full data, as shown in Figure \ref{fig:tasic-visual}. The visualisations of subsampled data with the optimal hyperparameters is given in the left column, and the middle column showcases the visualisation for the full data with the corresponding optimal normalized hyperparameters. 
Comparing the visualization using our method to choose perplexity and the default perplexity, we highlight that our method is better able to cluster like-data points. For example, the red clusters in our visualization have a linear structure but appear close together where using the default perplexity, the red clusters have a linear structure but do not appear together in the visualization.

\section{Conclusion}
We present an efficient and robust framework for selecting hyperparameters for DR methods and use DR techniques for data visualization such as t-SNE and UMAP to demonstrate the efficacy of our method. This work extends the applicability of previous works \citep{vu2021constraint} on this topic to large-scale data by leveraging sub-sampling and normalization of parameter spaces. We carry out both multi-objective tuning and sensitivity analyses for DR, addressing questions such as how much the sampling scheme affects the quality of DR and the trade-offs between computational complexity and performance. 

Considering $\mathbb{E}F(\gamma)$ instead of $F(\gamma)$ improves stability and robustness, 
we leace for future work level-set estimates (instead of maximum of metrics) to gain more robustness and graph-theoretic guarantees \citep{wilkinson2005graph} (e.g., Scagnostics-preserving subsampling in our pipeline \citep{wilkinson2008scagnostics}), answering the question of hyperparameter selection for large datasets raised in \citet{gove2022new}. Subsampling on the normalized parameter space enhances efficiency in evaluating metrics. 
We have not observed significant variations among different sampling regimes on the datasets and DR methods we investigated, but for highly structured \citep{hie2019geometric} or ultra-high-dimensional (i.e., $n<d$ for micro-arrays \citep{wilkinson_distance-preserving_2022}) DR tasks, we expect non-trivial modification of surrogate-based tuning methods need to be developed. 
{
\small
\bibliographystyle{chicago}
\bibliography{DRTuning}
}
\newpage

\newcounter{preappendixpage}
\setcounter{preappendixpage}{\value{page}}
\appendix
\renewcommand\thesection{\Alph{section}}
\renewcommand\thesubsection{\thesection.\Alph{subsection}}
{\Large\textbf{Supplementary Materials}}

Our code and data will be released at \url{http://github.com/} for reproducibility and production purposes.
\section{Efficient and Robust Hyperparameter Tuning Algorithm}
The proposed algorithm is a two-stage optimization process designed to find the optimal hyperparameters of a dimensionality reduction (DR) method. It operates on an input data matrix with $n$ samples and $d$ dimensions, with the goal of reducing the dimensionality to $d'$ while minimizing a performance metric $\tau$. The framework takes as input a DR method (denoted as $\mathcal{D}$), its hyperparameter space $\bm{\Gamma}$, a surrogate model $\mathcal{S}$, and a specified budget for pilot and sampling iterations.

The algorithm begins with an initialization step, taking a subsample of the data  (or if $n$ is small enough, the full dataset) from the input data matrix. In the pilot loop stage, the algorithm iterates through the pilot budget $N_1$, performing the DR method with randomly chosen hyperparameters from the space $\bm{\Gamma}$ on the subsampled dataset. It then computes the performance metric on the reduced dataset and appends the results to the corresponding sets.

Subsequently, the surrogate loop stage iterates through the sampling budget $N_2$ and fits the surrogate model with the collected hyperparameters and performance metrics. The next candidate hyperparameters are determined by optimizing the acquisition function defined by the surrogate model. The DR method is performed with these candidate hyperparameters on the subsampled dataset, and the performance metric is computed. Depending on the variability of the metric, the algorithm returns the mean or the mean adjusted by a multiple of variance as the next objective. The candidate hyperparameters and performance metrics are then appended to their respective sets.

Upon completion of the specified budget of pilot and sampling iterations, the algorithm returns the optimal set of hyperparameters corresponding to the best performance metric. The underlying premise is that the optimal hyperparameters found on the subsampled dataset will also be effective for the full dataset.

In Table \ref{tab:The-performance-metric}, we present several more examples of performance metrics that can fit into our framework. 
Mutual information 
, Kullback-Leibler (KL) divergence, and Area Under the Curve (AUC) with the $R_{NX}(K)$ 
curve, focus on quantifying the amount of information preserved or lost during the process of dimension reduction. For instance, mutual information indices measure the shared information between the original high-dimensional data and its reduced representation. KL divergence, on the other hand, quantifies the divergence between the probability distributions of the original and the reduced data, essentially measuring the information lost. The AUC under the $R_{NX}(K)$ curve is a measure of the quality of the dimension reduction technique,
with a higher AUC indicating better preservation of the data structure. 

\begin{table}[ht!]
\vspace{2mm}
\centering
\begin{tabular}{ccc}
\toprule 
\multirow{4}{*}{Task-independent} & Pearson correlation coefficient & based on the pairwise distances\tabularnewline
 & (between pairwise distance vectors) & between $\bm{X},\bm{X}^{*}$\tabularnewline
\cmidrule{2-3} \cmidrule{3-3} 
 & coranking measures \tablefootnote{Coranking scores are introduced by \cite{lee2009quality} and implemented
in $\mathtt{R}$ by \cite{kraemer2018dimred}.} & based on the ranks of pairwise\tabularnewline
 & (e.g., AUC, Q-global, Q-local) & distances between $\bm{X},\bm{X}^{*}$\tabularnewline
\midrule 
\multirow{4}{*}{Task-dependent} & \multirow{2}{*}{mis-classification rates} & for classification on $\bm{X}^{*}$\tabularnewline
 &  & using labels defined on $\bm{X}$\tabularnewline
\cmidrule{2-3} \cmidrule{3-3} 
 & clustering scores metrics\tablefootnote{We refer to \cite{ben2008measures} for detailed definition for clustering
scores.} & for clustering on $\bm{X}^{*}$\tabularnewline
 & (e.g., mutual information indices) & using features of $\bm{X}^{*}$\tabularnewline
\bottomrule
\end{tabular}

\caption{\label{tab:The-performance-metric}The performance metric considered
for evaluating DR methods for hyperparameter selection. }
\end{table}

\noindent 
\begin{algorithm}[H]
\label{alg:Subsampling-surrogate-based-hype}
\label{supp:algorithm}

\rule[0.5ex]{1\columnwidth}{1pt}
\SetAlgoLined

\textbf{Input:} $n\times d$ data matrix $\bm{X}$, subsample size $n'$, reduced dimension $d'$, DR method $\mathcal{D}$ and associated hyperparameter space $\gamma\in\bm{\Gamma}$, performance metric $\tau$, surrogate model $\mathcal{S}$, pilot budget $N_{1}$, sampling budget $N_{2}$ (usually $N_{1}\ll N_{2}$), (optional) the number of repeats $N_{\text{repeat}}$.\\
\KwResult{Optimal hyperparameter $\hat{\gamma}\in\bm{\Gamma}$.}
\textbf{Initialize:}\\
$\Gamma_{1}=[~]$, $T_{1}=[~]$\\
$\bm{X}^{\sharp}\in\mathbb{R}^{n'\times d}$: Subsample of $\bm{X}$ (or the full dataset if $n'=n$) \\
\For{$\ell$ in $1:N_{1}$}{
  Randomly select hyperparameters $\gamma^{\sharp}\in\bm{\Gamma}$\\
  $\bm{X}^{*}=\mathcal{D}_{\gamma^{\sharp}}(\bm{X}^{\sharp})$\\
  Compute $t^{\sharp}=\tau\left(\bm{X}^{\sharp},\bm{X}^{*}\right)$\\
  Append $\gamma^{\sharp}$ to $\Gamma_{\ell}$ to get $\Gamma_{\ell+1}$; and append $t^{\sharp}$ to $T_{\ell}$ to get $T_{\ell+1}$\\
}

\For{$m$ in $(N_{1}+1):(N_{1}+N_{2})$}{
  Fit $\mathcal{S}$ using $\Gamma_{m}$ as inputs and $T_{m}$ as responses\\
  Optimize the acquisition functions of $\mathcal{S}$ to get new $\gamma^{\sharp}$\\
  \tcp{Apply $\mathcal{D}_{\gamma^{\sharp}}$ to $\bm{X}^{\sharp}$ to get $\bm{X}^{*}$ under different initializations, and take sample means of metrics.}
  $T=[~]$\\
  \For{$i$ in $1:N_{\text{repeat}}$}{
  Initialize with a different random seed each time and compute $\bm{X}^{*}=\mathcal{D}_{\gamma^{\sharp}}(\bm{X}^{\sharp})$\\
  Compute $t^{\sharp}=\tau\left(\bm{X}^{\sharp},\bm{X}^{*}\right)$\\
 Store $t^{\sharp}$ in list $T$ \\
  
}
  Compute the mean $\widehat{\mathbb{E}}T$ of list $T$ as the next objective\\
    Append $\gamma^{\sharp}$ to $\Gamma_{m}$ to get $\Gamma_{m+1}$; and append $\widehat{\mathbb{E}}T$ to $T_{m}$ to get $T_{m+1}$\\ 
  }
\rule[0.5ex]{1\columnwidth}{1pt}
\caption{Subsampling Surrogated-based Hyperparameter Tuning Algorithm.}
\end{algorithm}

This algorithm shares similarities with regular BO, as both methods employ a surrogate model to approximate the objective function and use an acquisition function to guide the search for optimal hyperparameters. However, there are some novel aspects of this algorithm that differentiate it from standard BO:
\begin{itemize}
    \item Subsampling: The algorithm focuses on finding the optimal hyperparameters using a subsample of the dataset, aiming to maintain efficiency in evaluating the metrics. This is particularly beneficial when dealing with large datasets, as it significantly reduces computational costs.

    \item Dimension reduction: The algorithm is tailored for optimizing hyperparameters of dimensionality reduction (DR) methods, which is a specific application that requires unique consideration of the performance metrics and DR goals.

    \item Normalization and adaptability: The algorithm normalizes the hyperparameter spaces to ensure that the DR methods have the same parameter spaces for both the subsample and the full sample. This is a crucial aspect, as it allows the algorithm to maintain consistency between the subsample and the full sample, ensuring that the optimal hyperparameters found on the subsample generalize well to the full dataset. Furthermore, this normalization process allows the algorithm to be adapted to work with various DR methods, performance metrics, or dimensionality reduction goals, depending on specific applications or dataset characteristics, making it more versatile in handling different problems.

    \item Task-independent and dependent metrics: The algorithm considers the expected performance metric $\mathbb{E}F(\gamma)$ instead of the actual metric $F(\gamma)$, which increases its stability, robustness, and applicability to various task-independent and dependent metrics.
\end{itemize}
While the algorithm builds upon the foundation of BO, these novel aspects make it more suitable for optimizing hyperparameters in dimensionality reduction tasks and provide unique advantages over regular BO in this context.

The algorithm could also include  a (manual or automatic) procedure for determining the next objective based on the variability of specific metrics. These metrics are calculated by running the dimension reduction (DR) method multiple times with different initializations and computing a score $t^{\sharp}=\tau\left(\bm{X}^{\sharp},\bm{X}^{*}\right)$ for each run. The aim here is to capture a representative value that indicates the quality of the DR with the current hyperparameters.

For metrics that exhibit low variability (smaller variance during $N_{\text{repeat}}$ repeats), the algorithm returns the mean of the computed scores as the next objective. This is because, in such cases, the mean is a reliable measure of central tendency.

However, for metrics with high variability (larger variance during $N_{\text{repeat}}$ repeats), the empirical mean might not be an adequate representation due to the influence of potential outliers. In such cases, the algorithm proposes two alternatives. The first one is to return a specific quantile (for example, the median) of the computed scores as the next objective. The median is a robust measure of central tendency unaffected by outliers. The second alternative is to return the mean of the computed scores minus a multiple of the variance. This adjustment accounts for the variability of the metrics and makes the selected next objective more conservative. The choice between these two alternatives could be based on the specific characteristics of the dataset and the DR method in use.

\section{Visualization for COIL Dataset}\label{SM:Visual_COIL}
\begin{figure}[ht]

\centering
\begin{subfigure}[t]{0.375\textwidth}
\centering
\includegraphics[width=\linewidth]{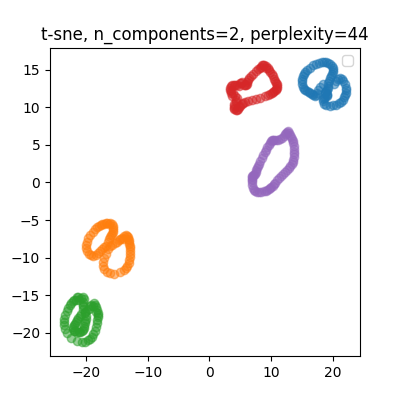}

\end{subfigure}
\begin{subfigure}[t]{0.375\textwidth}
\includegraphics[width=\linewidth]{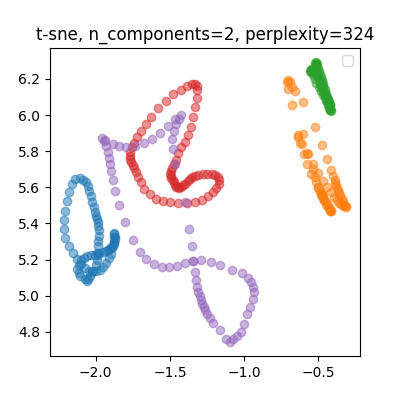}

\end{subfigure}
\begin{subfigure}[t]{0.15\textwidth}
\includegraphics[width=\linewidth]{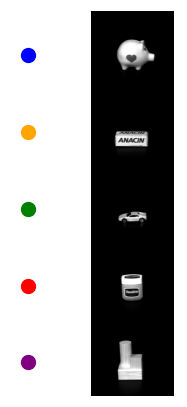}
\end{subfigure}
 \caption{2-dimensional visualization for $5$ of the objects from the COIL dataset using t-SNE with different perplexities.}\label{fig:coil5}
\end{figure}

In Figure \ref{fig:coil5}, we compare the performance of t-SNE when different metrics are minimized. 
When the perplexity is set to 44 (normalized perplexity 0.19), we observe a distinct separation of the data points from different classes,
and at this perplexity both $1-\text{AUC}$ and classification error rate are minimized. Moreover, the metric $1-\text{Q-global}$ attains its minimum when the perplexity is set to 324 (normalized perplexity 0.9), indicating t-SNE better preserves global structure.

\section{Surrogate-based Sensitivity Analysis for UMAP}\label{sec:SA}
When there is more than one hyperparameter to select, such as in UMAP, we can perform surrogate-based sensitivity analyses for hyperparameters with respect to the chosen performance metric (i.e., Sobol and Shapley analysis \citep{owen2014sobol}).
The Sobol sensitivity analysis conducted on the given fitted surrogate
model for $\mathbb{E}F(\gamma)$ provides valuable insights into the
influence of the hyperparameters (number of nearest neighbors (n\_neighbor)
and minimum distance (min\_distance)) on the variance of the resulting
chosen performance metric. A comparison of the first-order ($S_1$) and
total-order ($ST$) sensitivity indices reveals distinct effects of the
parameters on the model output (i.e., performance metric) of the chosen
metric $1-\text{AUC}$. This analysis helps us understand how each hyperparameter
affects the chosen performance metric. 
\begin{table}[h!]
\centering %
\vspace{1em}
\begin{tabular}{c|c|c|c|c}
\hline 
Parameter  & $S_{1}$  & $S_{1}$ Conf.  & $ST$  & $ST$ Conf.\tabularnewline
\hline 
n\_neighbor  & 4.89e-04  & 4.70e-04  & 1.00049  & 9.61e-01\tabularnewline
min\_distance  & 4.89e-04  & 9.29e+24  & 1.00049  & 1.22e+25\tabularnewline
\hline 
\end{tabular}
\vspace{1em}
\caption{Sobol sensitivity indices based on surrogate for the tuning samples
$(\gamma,F(\gamma))$ and their confidence intervals for the UMAP
experiment on the unbalanced 2-cluster dataset.}
\label{table:sobol_results_new} 
\end{table}

In Table \ref{table:sobol_results_new} for the unbalanced 2-cluster
dataset, the first-order sensitivity indices ($S_{1}$) and their
respective confidence intervals ($S_{1}$ Conf.) are presented, indicating
that the individual contribution of each input parameter to the performance
metric variance. The total-order sensitivity indices ($ST$) and their
confidence intervals ($ST$ Conf.) are displayed, representing the
combined effect of each parameter, accounting for both its individual
contribution and its interactions with other parameters.

Both parameters have similar $S_{1}$ and $ST$ values, suggesting
that they have comparable individual and overall impacts on the performance
metric variance. However, the confidence intervals for min\_distance
are significantly larger, particularly for $S_{1}$, which indicates
a higher uncertainty in the sensitivity index estimates for min\_distance.
This is intuitive since both hyperparameters can affect the quality
of DR on unbalanced clusters.

\begin{table}[h!]
\centering%
\vspace{1em}
\begin{tabular}{c}
Full\tabularnewline
\hline 
\begin{tabular}{c|c|c|c|c}
\hline 
Parameter  & $S_{1}$  & $S_{1}$ Conf.  & $ST$  & $ST$ Conf.\tabularnewline
\hline 
n\_neighbor  & 0.005881  & 0.002237  & 1.003019  & 0.002075\tabularnewline
min\_distance  & 0.000307  & 0.019511  & 1.896585  & 3.475031\tabularnewline
\hline 
\end{tabular}\tabularnewline
\hline 
Subsampled (Uniform)\tabularnewline
\hline 
\begin{tabular}{c|c|c|c|c}
\hline 
Parameter  & $S_{1}$  & $S_{1}$ Conf.  & $ST$  & $ST$ Conf.\tabularnewline
\hline 
n\_neighbor  & 0.013529  & 0.001716  & 1.008246  & 0.004326\tabularnewline
min\_distance  & 0.004862  & 0.007099  & 1.510901  & 0.429790\tabularnewline
\hline 
\end{tabular}\tabularnewline
\hline 
\end{tabular}
\vspace{1em}
\caption{Sobol sensitivity indices based on surrogate for the tuning samples
$(\gamma,F(\gamma))$ and their confidence intervals for the UMAP
experiment on the COIL dataset.}
\label{table:sobol_results} 
\end{table}

Table \ref{table:sobol_results} indicates that the individual contribution
of each input parameter to the performance metric variance. In both
analyses with full or subsample during tuning, n\_neighbor has a higher
first-order sensitivity index  
($S1$) than min\_distance, indicating
that n\_neighbor has a larger individual contribution to the metric
variance. The total-order sensitivity index ($ST$) values for both
n\_neighbor and min\_distance show differences between the two analyses,
partially due to the increased uncertainty introduced by subsampling.
In analysis based on full sample (no subsampling during tuning), min\_distance
has a much larger $ST$ value, suggesting that min\_distance has a
more substantial overall impact on the metric variance when accounting
for interactions between these two parameters on this COIL dataset,
which is supported by the analysis on subsampled dataest as well. 

\section{Sine Dataset}\label{supp:sine dataset}
This is a small-scale, synthetic dataset that allows us to investigate the framework of data with underlying geometric structures. The full dataset $\{ x_i\}_{i=1}^{n} \in \mathbb{R}^4$ where the vector $x_i = [z(i),\sin(z(i)),\sin(2z(i)),\sin(3z(i))]$, were $z(i) =  \frac{2\pi}{n-1}(i-1) - \pi $. In other words, the dataset is the set of tuples $(z,\sin(z),\sin(2z),\sin(3z))$ evaluated on $n=200$ equally spaced points $z \in [-\pi, \pi]$.

Since our tuning is surrogate-based, we can also apply other supervised model analysis tools like Shapely values \citep{owen2014sobol} or LIME \citep{ribeiro2016should}.

\begin{figure}[t]
\centering
\includegraphics[width=1\linewidth]{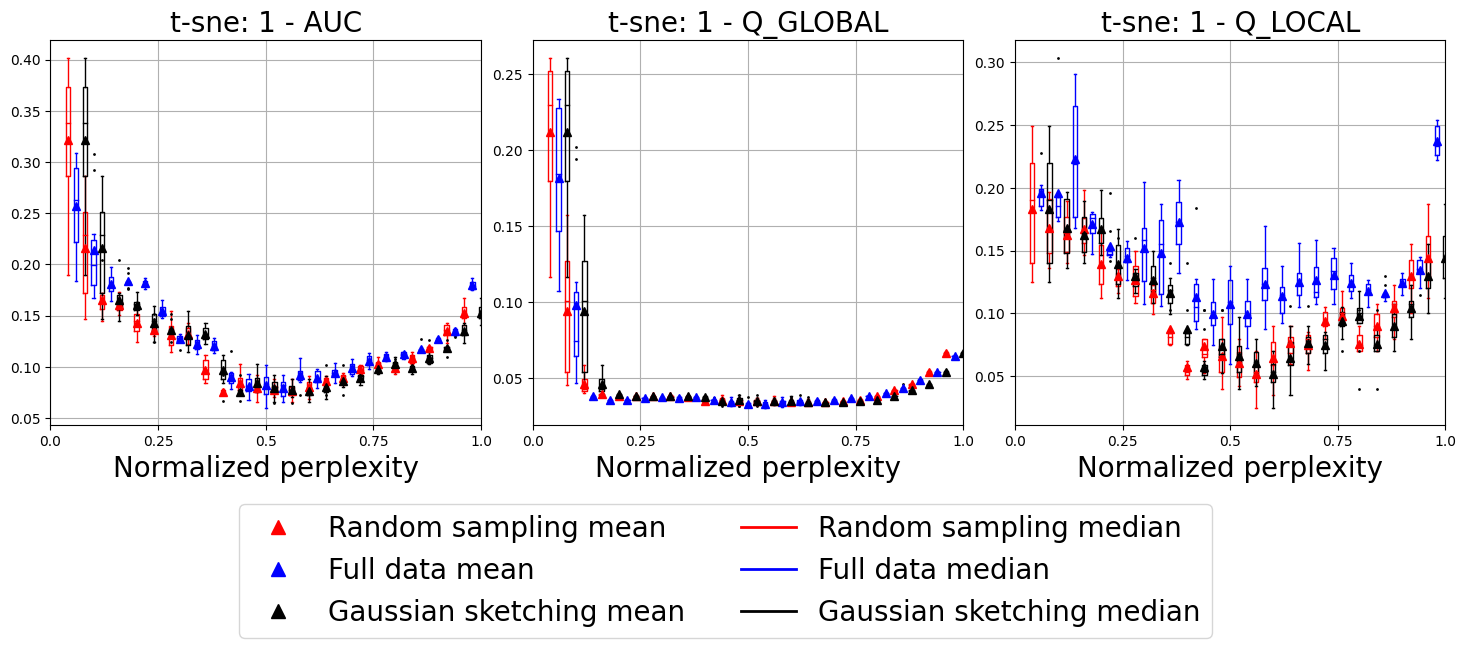}
 \caption{Boxplots for the performance of t-SNE for the Sine Dataset.}
\label{fig:both} 
\end{figure}
For our next experiment, we apply our framework to the Sine dataset. In this experiment, we consider the following metrics, AUC (for local structure), Q-global (for global structure), and the error rate for classification performance.

In Figure \ref{fig:both}, we observed that the choice of subsampling regime can influence the discrepancy between the subsample-based metric $\mathcal{S}_{\gamma}(\bm{X}^\sharp)$ and the true performance metric function $\mathcal{S}_{\gamma}(\bm{X})$. Random subsampling, for instance, may not capture the full diversity of the data, leading to ignorance in local details. Additionally, the selected performance metric plays a crucial role. If it doesn't adequately encapsulate aspects of the data sensitive to the hyperparameters, the resulting function may deviate from the true relationship.
The complexity and structure of the full dataset pose challenges to the subsampling \citep{hie2019geometric} and surrogate modeling \citep{luo2022spherical}, possibly causing it to miss key interaction effects, leading to a divergence from the true function form. Additionally, the choice of surrogate model, optimization procedure, and inherent data variability are crucial in reducing the randomness. 
For instance, an inflexible surrogate model might fail to capture the true relationship, and an inadequate optimization procedure might not fully explore the parameter space. High data heterogeneity could further influence the function shape derived from a subsample, which makes it more important to examine multiple metrics simultaneously as our framework suggests.  

From Figure \ref{fig:both}, the optimal perplexity for the full data, according to average ratio, is around 160 (normalized perplexity =0.8), while that for the subsampled data is near 50 (normalized perplexity =0.8). In contrast, according to AUC, Q-global, or Q-local, the optimal perplexity for the original data is close to 100 (normalized perplexity = 0.5), and that for the subsampled data is close to 33 (normalized perplexity = 0.5).
Under the normalized perplexity, the performance of the original and subsampled data is similar. This suggests that using the same normalized perplexity for t-SNE on the subsampled data yields results comparable to applying t-SNE on the full data and then subsampling the reduced data.

\begin{figure}[t]
\centering \begin{subfigure}[t]{0.7\linewidth} \includegraphics[width=1\linewidth]{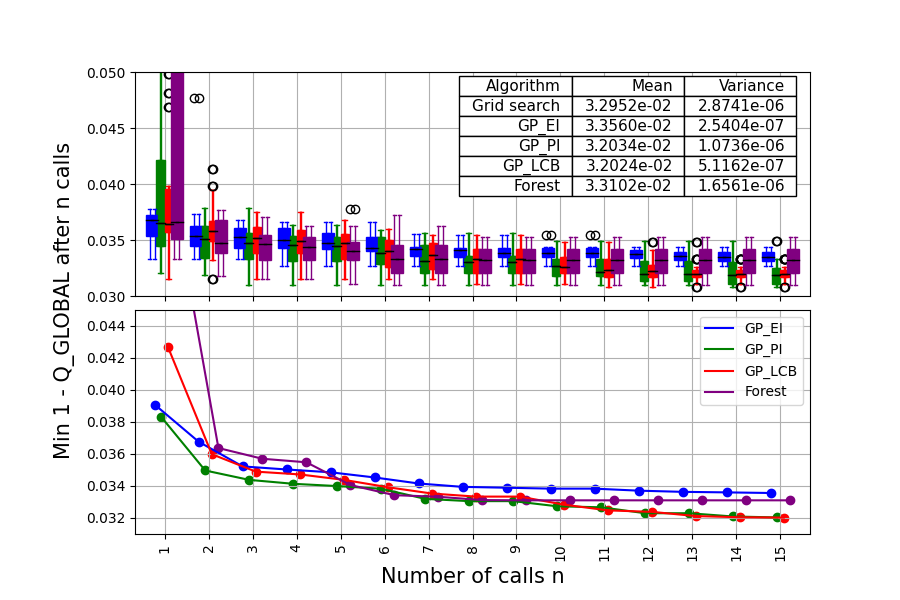}
\end{subfigure} \quad{}\begin{subfigure}[b]{0.7\linewidth} \includegraphics[width=1\linewidth]{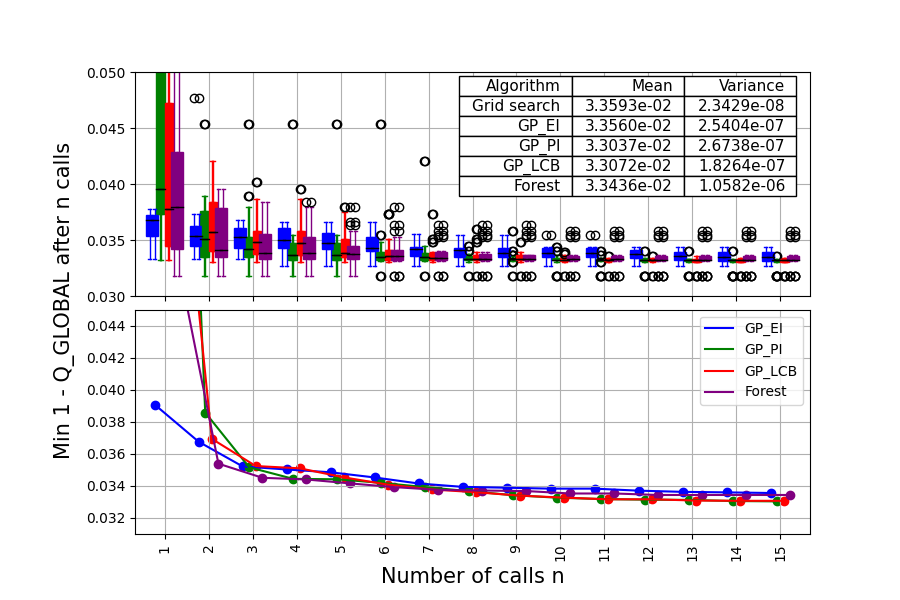}
\end{subfigure} \caption{Convergence plot for searching optimal perplexities for t-SNE for
the 4-dimensional vector $[x,\sin(x),\sin(2x),\sin(3x)]$.  The results for original data is on the top and those for the subsampled data is on the bottom.}
\label{fig:single_coil}
\end{figure}

Finally, we include the surrogate-based tuning for both original data
and the subsampled data. In contrast with Example 1, where we use the metric AUC, in this example we choose a different metric Q-global which has a much flatter landscape as shown in Figure \ref{fig:both}. We use GP-EI for the surrogate-based tuning and assign the budget $N_0=N_1+N_2=5+15$ and run the training for $20$ times. The convergence plots for both cases are given in Figure \ref{fig:single_coil}. We see from both plots that though the landscape for Q-global is flat, the surrogate-based tuning is able to converge to the minimum in a few iterations and in order to find the optimal perplexity for the original data, we could instead more efficiently tune on the subsampled data.

\section{Vehicle Dataset}\label{supp:vehicle dataset}
We consider the vehicle dataset from the LIBSVM repository. The Vehicle dataset consists of features extracted from the silhouettes of vehicles, consisting of 4 classes and 846 data with 18 features for each data. The objective is to classify a given silhouette as one of four types of vehicles: 'Opel', 'Saab', 'Bus', or 'Van'. The features measure various properties of the silhouette, such as compactness, circularity, and distance between the center of the vehicle and a line through the center of the vehicle and the rear axle. 

\begin{figure}[ht!]
\centering
\includegraphics[width=.75\linewidth]{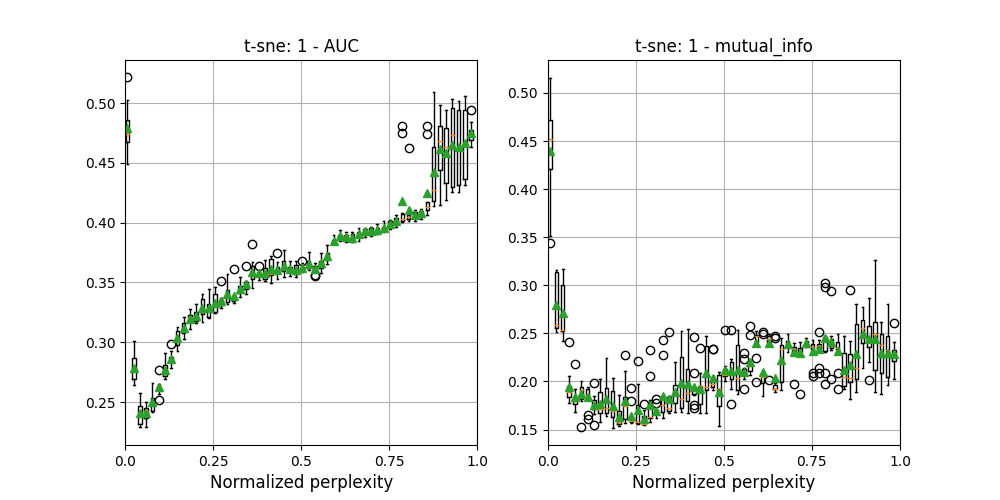}
 \caption{Boxplots for the performance of t-SNE for the vehicle dataset using AUC and mutual information for all possible  normalized perplexities, where the medians are marked by green triangles}
\label{fig:vehicle_sample}
\end{figure}

We sample a third from the full data and plotted the performance metrics at each perplexity in Figure \ref{fig:vehicle_sample}. We next include surrogate-based tuning using AUC on the subsampled data as a proxy for the full data. The optimal normalized perplexity obtained in the tuning for the subsampled data is 0.06. We then perform t-SNE on the full dataset with perplexity $51 \approx 846\times 0.06$. In this example, we are able to find the optimal perplexity under AUC. However, without preprocessing the dataset, we are not able to use t-SNE to distinguish different classes of data. A similar observation can be found in Figure 3 of \citep{Vehicle2009}.

\begin{figure}[ht!]
\centering \begin{subfigure}[b]{0.5\linewidth} \includegraphics[width=1\linewidth]{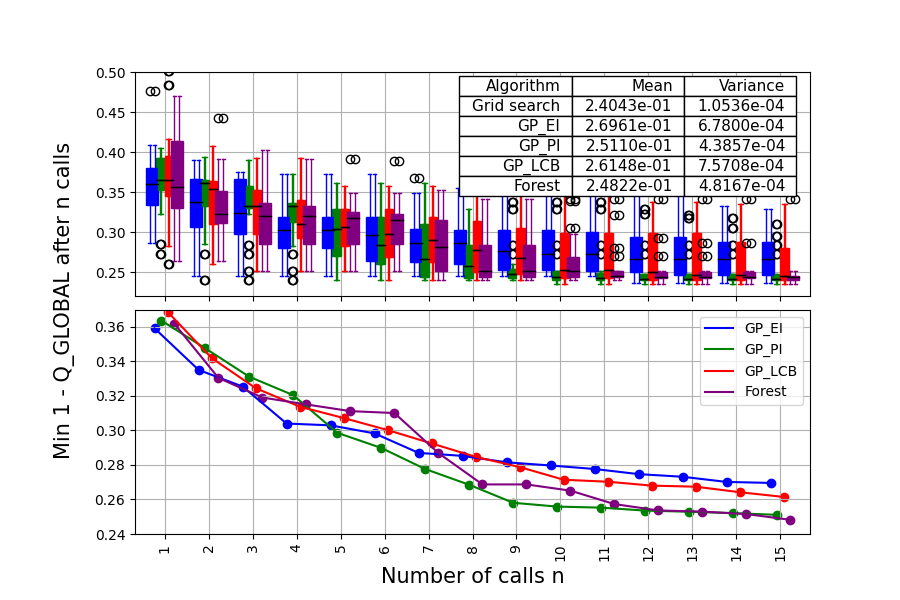}
\caption{Convergence plot for searching optimal perplexities for t-SNE}
\end{subfigure}
\quad{}
\begin{subfigure}[b]{0.35\linewidth} \includegraphics[width=1\linewidth]{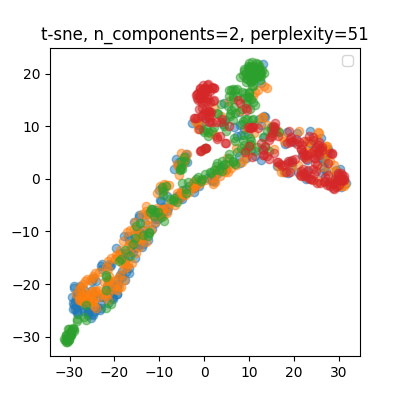}
\caption{Visualization of t-SNE at the optimal perplexity 
}
\end{subfigure} 
\caption{Surrogate-based tuning for the subsampled dataset and the corresponding visualization on the full dataset.}
\label{fig:vehicle_gp}
\end{figure}

Additionally, the convergence plot Figure \ref{fig:vehicle_gp} shows typical behavior of ``flatness'' when the performance metric landscape is flat. As future work, we may want to include early-stopping techniques from BO \citep{swersky2014freeze} to terminate the hyperparameter tuning. On one hand, this saves the computational cost for further evaluations; on the other hand, this could serve as a flag for potential inappropriateness for applying certain DR method on the given dataset. 

\section{Reuters Dataset}
\begin{figure}[ht!]
    \centering
    \includegraphics[width=\linewidth]
    {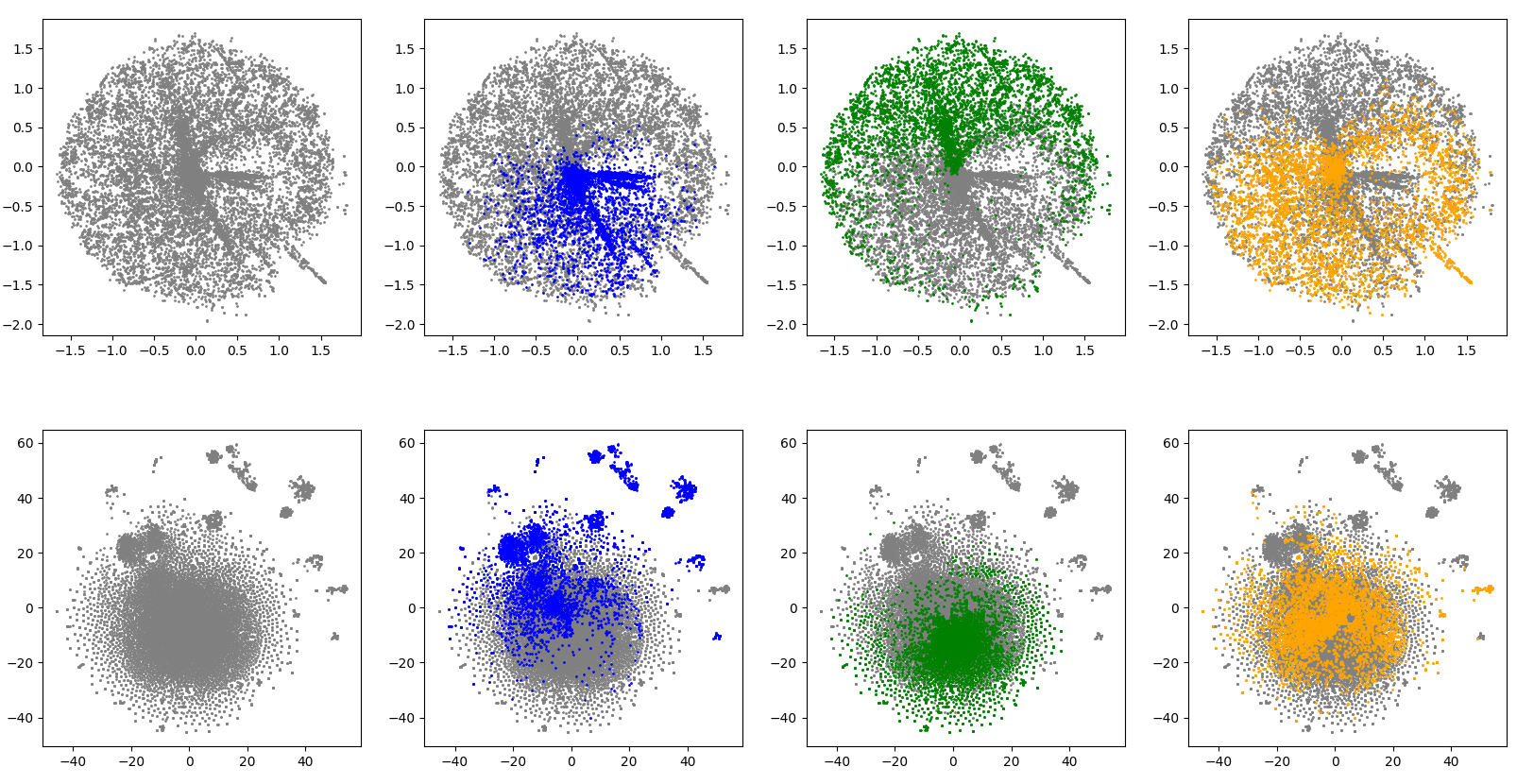}
    \caption{Visualization of the Reuters dataset with individual classes highlighted. The first column shows the visualizations without any class labels, which are oftentimes unknown in the data exploratory phase. The second, third, and fourth columns highlight different classes in this dataset, which corresponds to C15 (performance), GCAT (government and social), and CCAT (corporate and industrial), respectively. The top row uses perplexity = 6,418 (normalized perplexity = 0.4500) and the bottom row uses the default perplexity of 30 (normalized perplexity = 0.0021). }
    \label{fig:reuters-overlaps}
\end{figure}

In Figure~\ref{fig:reuters-overlaps}, we highlight that the visualization produced via our framework provides a more informative view of the data. In this figure, we highlight individual classes in the Reuters dataset. 

We see that the C15 and CCAT  are better separated from the GCAT when the perplexity is chosen using the surrogate model compared to the default t-SNE perplexity parameter, which is a reasonable expectation as C15 can be considered a subclass of CCAT in the Reuters dataset. 
Moreover, at the optimal perplexity, the visualization performs better in classification and clustering tasks, as demonstrated by the evaluation metrics, namely classification error rate (0.3102 vs 0.3242) and NMI (0.2429 vs 0.2054).

\end{document}